%% file: preprint.tex
\documentclass{article} %
\usepackage{iclr2027_conference,times}
\usepackage[utf8]{inputenc}
\usepackage[T1]{fontenc}
\usepackage{hyperref}
\usepackage{url}
\usepackage{booktabs}
\usepackage{amsfonts}
\usepackage{nicefrac}
\usepackage{microtype}
\usepackage{xcolor}
\usepackage{amsmath}
\usepackage{graphicx}
\usepackage{multirow}
\usepackage{enumitem}
\usepackage{algorithm}
\usepackage{algpseudocode}
\usepackage{amssymb}
\usepackage{tikz}
\usepackage{xspace}
\newcommand{\xhdr}[1]{\noindent{{\bf #1.}}}

\usepackage{caption}
\usepackage{subcaption}
\usepackage{float}
\usepackage{caption}
\usepackage{threeparttable}
\usepackage{titlesec}
\usepackage{wrapfig}
\usepackage{amsmath}
\usepackage{amssymb}
\titlespacing*{\paragraph}{0pt}{0pt}{0.5em}
\definecolor{modelcolor}{HTML}{F0F0FF}
\usepackage{tcolorbox}
\usepackage{tabularx}
\usepackage{array}
\usepackage{ragged2e}
\usepackage{booktabs}
\usepackage{hyperref}
\usepackage{url}
\usepackage{caption}
\usepackage{makecell}
\tcbset{
    boxsep=0pt,
    left=3pt,
    right=3pt,
    top=3pt,
    bottom=3pt,
    before skip=3pt,
    after skip=3pt,
    boxrule=1.5pt,
colback=blue!6!white,
colframe=blue!20!gray
}

\usepackage{xcolor}

\input{math_commands.tex}

\usepackage{hyperref}
\usepackage{url}

\title{When Do Biological Reasoning Models Use Their Biological Inputs?}

\author{%
  Ada Fang\\
  Harvard University\\
  \texttt{ada\_fang@g.harvard.edu} \\
  \And
  Nikitha Thoduguli\\
  Massachusetts Institute of Technology\\
  \texttt{tnikitha@mit.edu} \\
  \And
  Lukas Fesser\\
  Harvard University\\
  \texttt{lukas\_fesser@g.harvard.edu} \\
  \And
  Hanlin Zhang\\
  Harvard University\\
  \texttt{hanlinzhang@g.harvard.edu} \\
  \And
  Sham Kakade\\
  Harvard University\\
  \texttt{sham@g.harvard.edu} \\
  \And
  Marinka Zitnik \\
  Harvard University\\
  \texttt{marinka@hms.harvard.edu} \\
}

\iclrfinalcopy %
\begin{document}

\maketitle
\fancyhead{}
\fancyfoot{}
\cfoot{\thepage}

\begin{abstract}
Biological reasoning models use post-training to connect LLMs to biological inputs, typically representations from a biological foundation model and biological text such as gene and pathway descriptions, functional annotations, and gene lists. Their benchmark accuracy is taken as evidence that the LLM reasons over their biological inputs.
We test this assumption in six biological reasoning models across DNA, protein, and single cell tasks. We perturb one biological input while holding the query and other inputs fixed, construct evidence conflicts that pair the foundation model representation of one genome, protein, or cell with the text of another, fit linear probes to the representations the language model receives, and analyze reasoning traces against the biological inputs.
We find that current post-training strategies do not ensure that foundation model representations contribute to task performance. Evo2 and ESM3 contribute little to BioReason and BioReason-Pro performance on the evaluated tasks. Shuffling the DNA sequence barely changes BioReason disease prediction accuracy, and in evidence conflicts the two models follow the text in 97.9\% and 99.7\% of cases. Linear probes trained on the Evo2 and ESM3 representations predict the task targets, so these foundation models encode information relevant to the task, but provide limited overall performance improvement to BioReason and BioReason-Pro.
In contrast, foundation model inputs contribute to ChatNT, Prot2Text-V2, and CellWhisperer performance, and differentially expressed genes in the gene sentence contribute to Cell2Sentence-Scale performance.
Across SFT and RL checkpoints of BioReason-Pro and 42 BioReason checkpoints, increases in accuracy do not imply greater performance contributions from biological inputs.
In reasoning traces, BioReason traces misstate nucleotide changes, while BioReason-Pro traces describe functions omitted from final predictions under evidence conflicts.
Future work should test whether post-training objectives that reward correct use of biological inputs improve their contribution to task performance.

\end{abstract}

\begin{center}
Website: \href{https://zitniklab.hms.harvard.edu/biomirage}
{\textcolor{magenta}{\nolinkurl{https://zitniklab.hms.harvard.edu/biomirage}}}

Code: \href{https://github.com/mims-harvard/bio-mirage}
{\textcolor{magenta}{\nolinkurl{https://github.com/mims-harvard/bio-mirage}}}
\end{center}

\section{Introduction}
\input{iclr2026/010introduction}

\vspace{-3mm}
\input{iclr2026/011related_work}
\vspace{-3mm}
\section{Approach}
\vspace{-3mm}
\input{iclr2026/020methods-problem-defn}

\vspace{-3mm}
\section{Results: Biological Input Use in Reasoning Models}
\input{iclr2026/031modality-dependence}

\input{iclr2026/032model-variants}

\input{iclr2026/033rationales}
\input{iclr2026/034post-training}

\section{Discussion}
\vspace{-3mm}

\input{iclr2026/040discussion}

\subsection*{Acknowledgments}
A.F., L.F., and H.Z. are supported by the Kempner Graduate Fellowship at Harvard University.
We gratefully acknowledge support, in part, by NSF CAREER 2339524, Harvard Data Science Initiative, Amazon Faculty Research, Google Research Scholar Program, AstraZeneca Research, Roche Alliance with Distinguished Scientists (ROADS) Program, Sanofi iDEA-iTECH Award, GlaxoSmithKline Award, Boehringer Ingelheim Award, Merck Award, Optum AI Research Collaboration Award, Pfizer Research, Gates Foundation (INV-079038), Chan Zuckerberg Initiative, John and Virginia Kaneb Fellowship at Harvard Medical School, Biswas Computational Biology Initiative in partnership with the Milken Institute, Harvard Medical School Dean’s Innovation Fund for the Use of Artificial Intelligence, and the Kempner Institute for the Study of Natural and Artificial Intelligence at Harvard University. This work was delivered as part of the Aurora project supported by the Cancer Grand Challenges partnership funded by Cancer Research UK (CGCAI1-Mar26/100001). Any opinions, findings, conclusions, or recommendations expressed in this material are those of the authors and do not necessarily reflect the views of the funders.
Any opinions, findings, conclusions or recommendations expressed in this material are those of the authors and do not necessarily reflect the views of the funders.

\bibliography{iclr2026/iclr2026_conference}
\bibliographystyle{iclr2027_conference}
\newpage
\appendix
\setcounter{figure}{0}
\setcounter{table}{0}
\renewcommand{\thefigure}{S\arabic{figure}}
\renewcommand{\thetable}{S\arabic{table}}

\renewcommand{\theHfigure}{S\arabic{figure}}
\renewcommand{\theHtable}{S\arabic{table}}

\input{iclr2026/091appendix_rq1}

\input{iclr2026/092appendix_rq2}

\input{iclr2026/093appendix_rq3}

\input{iclr2026/094appendix_rq4}

\end{document}

%% file: math_commands.tex
\usepackage{amsmath,amsfonts,bm}

\def\eqref#1{equation~\ref{#1}}

\def\1{\bm{1}}

\DeclareMathAlphabet{\mathsfit}{\encodingdefault}{\sfdefault}{m}{sl}
\SetMathAlphabet{\mathsfit}{bold}{\encodingdefault}{\sfdefault}{bx}{n}

%% file: iclr2026/010introduction.tex
Biological reasoning models combine language models with biological inputs
to make predictions about proteins, genomes, and cells. These inputs include
continuous representations from biological foundation models and biological
text, such as functional annotations, gene and pathway descriptions, and
expression-ranked gene lists. Recent examples include DNA--LLM models
\citep{fallahpour2025bioreason, chatnt}, single-cell transcriptomic language
models \citep{levine2024cell2sentence, cellwhisperer}, and protein reasoning
models \citep{procyon, bioreasonpro, liu-etal-2024-prott3, ProTrek}.
Benchmark performance alone, however, does not establish which biological
inputs drive prediction. Predictive annotations can support accurate answers
even when accompanying sequence or transcriptomic representations are
underused, so benchmark improvements need not reflect greater use of these
representations \citep{geirhos2020shortcut, degrave2021ai}.

Attributing a prediction to a particular biological input is difficult because these models often
receive multiple representations of the same protein, genome, or cell. These representations can
encode overlapping biological properties at different levels of abstraction. A limited performance contribution from one input may therefore reflect support from another input, limited predictive value of the evaluated input, or limited use of a predictive input by the reasoning model. Benchmark performance alone cannot distinguish among these alternative explanations.

\begin{figure}[t]
    \centering
    \vspace{-2mm}
    \includegraphics[width=\linewidth]{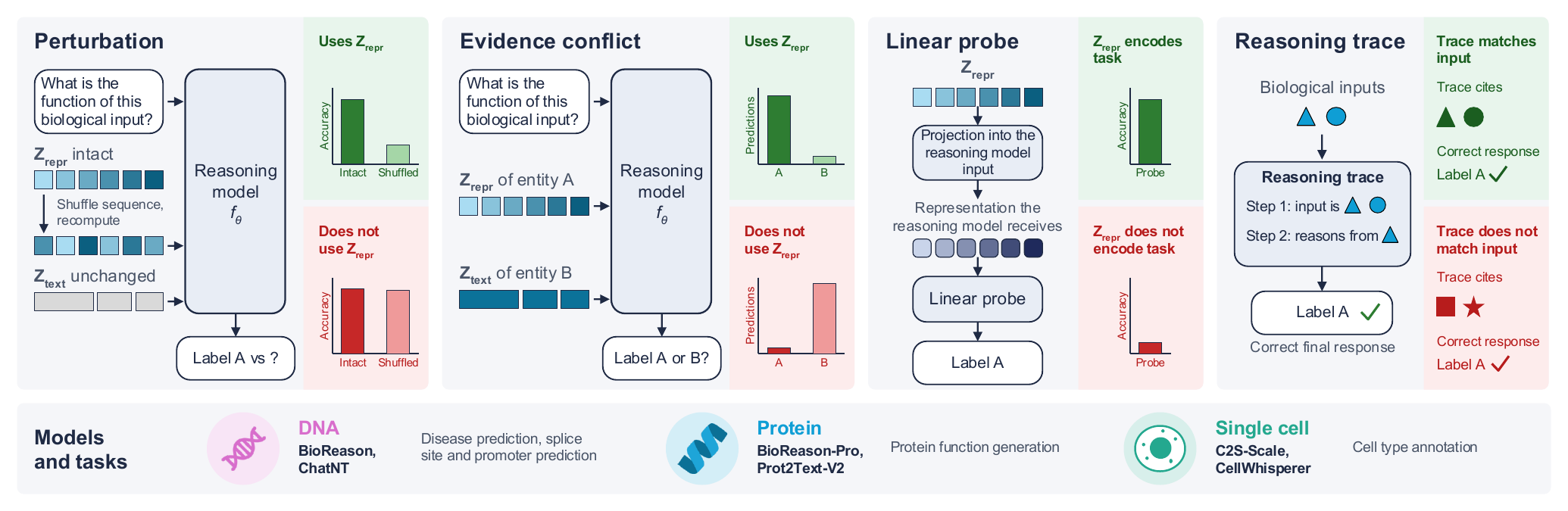}
    \vspace{-7mm}
    \caption{Evaluating biological input usage in biological reasoning models across DNA, proteins and single-cell.
    }
    \vspace{-4mm}
    \label{fig:fig1}
\end{figure}

Related problems have been studied in vision--language models, where predictions can exploit
language priors or remain insensitive to relevant visual content
\citep{goyal2017making, agrawal2018don, hessel2020does, asadi2026mirage}.
Representation probing has separately shown that a target can be predicted from a learned
representation without implying that the model uses that representation for its own
prediction \citep{ravichander-etal-2021-probing, elazar-etal-2021-amnesic}. Biological reasoning
models exacerbate these challenges because several representations derived from the same biological
object may be available to the model simultaneously. Existing evaluations of these models emphasize task performance, leaving unresolved which biological inputs
contribute to prediction.

\xhdr{Present work}
Biological reasoning models are multimodal LLMs that receive a user task together with biological inputs, such as foundation model representations of a genome, protein, or cell and biological text, and generate a response that contains an answer and often a reasoning trace (Figure~\ref{fig:fig1}). We develop tests that determine whether the response of such a model depends on each of its biological inputs. The first test (``Perturbation'') perturbs one biological input while holding the user task and the other inputs fixed and measures changes in the answer and task performance. The second test (``Evidence conflict'') places inputs from two different genomes, proteins, or cells in the same prompt so that they support different answers, and identifies the input the model follows. When an input contributes little to performance, linear probes on the representation the model receives determine whether that input encodes the task target, which separates an uninformative input from an informative input with a limited performance contribution (``Linear probe''). The fourth test (``Reasoning trace'') evaluates whether biological claims in generated traces and prompted rationales agree with the supplied inputs, respond to input perturbations, and remain consistent with final predictions. We apply these tests to six models (Table~\ref{tab:models_tasks}) across post-training checkpoints and model scales.
Our main findings are:

\begin{itemize}[leftmargin=*, labelindent=0pt, topsep=0pt, itemsep=0pt]
\item \textbf{Foundation model representations can contribute little to performance even when they encode the task target.}
BioReason and BioReason-Pro prefer the biological text in evidence conflicts, and Evo2 and ESM3 representations contribute little to their performance, although linear probes predict the task targets from these representations. Foundation model representations contribute to ChatNT, Prot2Text-V2, and CellWhisperer performance, and differentially expressed genes (DEGs) in the gene sentence contribute to C2S-Scale performance.
\item \textbf{Higher accuracy after post-training does not indicate a greater performance contribution from the foundation model.}
SFT and RL increase the accuracy of BioReason and BioReason-Pro without increasing the performance contribution from Evo2 and ESM3 representations, and we observe the same result across 42 BioReason checkpoints. Scaling C2S-Scale from 2B to 27B increases accuracy and the performance contribution from DEGs.

\item \textbf{Generated reasoning does not consistently connect biological evidence to final predictions.}
BioReason makes incorrect claims about nucleotide substitutions, while BioReason-Pro discusses functions absent from its final GO annotations under evidence conflicts. For C2S-Scale 27B, marker enrichment is concentrated in the first few gene references and is lost across the full rationale where it over-references ribosomal protein genes.

\item \textbf{Auxiliary supervision increases sequence sensitivity, but the edited-base prediction contributes little to disease prediction performance.}
Training BioReason to predict the edited bases makes this prediction depend on Evo2 representations. Shuffling the sequence lowers accuracy on the edited base pair by 0.549 and disease prediction accuracy by only 0.021. However, this improvement in predicting edited bases is limited to genomes seen during training.

\end{itemize}

%% file: iclr2026/011related_work.tex
\section{Related Work}
\vspace{-3mm}

\xhdr{Biological reasoning models}
Biological reasoning models differ in their input representations.
ChatNT uses Nucleotide Transformer representations, while BioReason
combines Evo2 with gene and pathway text
\citep{chatnt, evo2, fallahpour2025bioreason}.
Prot2Text-V2 combines ESM2 representations with protein name and taxon
text, whereas BioReason-Pro combines ESM3 with GO-GPT predictions
and InterPro annotations \citep{fei2025prottextv, esm3, bioreasonpro}.
Cell2Sentence-Scale represents transcriptomes as expression-ranked gene
sentences, while CellWhisperer uses continuous transcriptomic representations
\citep{levine2024cell2sentence, cellwhisperer}.
Other protein models connect biological representations with language
for retrieval, annotation, or generation
\citep{procyon, ProTrek, protst, liu-etal-2024-prott3,
jararweh-etal-2025-protein2text, wang-etal-2024-instructprotein,
char2025protnote}.
Evaluations largely emphasize task performance, scale, and post-training
\citep{biopostrain, wu-etal-2025-rethinking-text, ICLR2026_047c84ec,
he-etal-2025-biology, dip-etal-2026-llm4cell}.
We instead test whether predictions depend on biological input content,
distinguishing target predictability from model use.

\xhdr{Multimodal representation learning and evaluation}
Multimodal learning literature studies how representations from different modalities can
be aligned, combined, and decomposed into shared or modality-specific structure
\citep{liang2024multimodal, dufumier2025align, huh2024platonic,
gupta2026canonicalizing, zhang2026apollo, richter2026beyond}.
Complementary work evaluates whether multimodal models use the inputs provided to them.
Vision--language models can exploit language priors, remain insensitive to visual content, or
improve benchmark performance without increased cross-modal interaction
\citep{goyal2017making, agrawal2018don, NEURIPS2019_51d92be1,
hessel2020does, frank-etal-2021-vision, asadi2026mirage, gao2026do}.
Counterfactual inputs and controlled perturbations have therefore been used to test whether
predictions respond to the content of a particular modality
\citep{Niu_2021_CVPR, Thrush_2022_CVPR}.
Our work focuses on biological reasoning models, where multiple representations of the same
protein, genome, or cell may be available simultaneously, and asks which of these biological inputs
contribute to the downstream prediction.

\xhdr{Predictive representations and generated rationales}
Linear probes provide a simple test of
predictability from learned representations \citep{alain2017iclr-understanding, hewitt-liang-2019-designing}, while subsequent
work has shown that probe performance and downstream model behavior should be evaluated
separately \citep{ravichander-etal-2021-probing, elazar-etal-2021-amnesic}. Generated rationales
introduce a related distinction because plausible explanations need not identify the features that
affect model predictions
\citep{jacovi-goldberg-2020-towards, deyoung-etal-2020-eraser,
wiegreffe-etal-2021-measuring, NEURIPS2023_ed3fea90}. We use simple predictive baselines to
interpret limited performance contributions from a biological input and compare features that affect prediction with
those referenced in generated rationales.

%% file: iclr2026/020methods-problem-defn.tex
\subsection{Problem Formulation}
\label{sec:problem_formulation}
\vspace{-2mm}
Biological reasoning models take a text query \(q\), which specifies the task and response format, together with biological inputs
\(\mathbf{Z} = (\mathbf{Z}_{\mathrm{repr}}, \mathbf{Z}_{\mathrm{text}})\)
associated with a biological modality \(X\), such as protein sequence, genomic sequence,
or a single-cell transcriptome. We distinguish continuous biological representations
\(\mathbf{Z}_{\mathrm{repr}}\) from biological inputs represented as text
\(\mathbf{Z}_{\mathrm{text}}\). For target \(y\), the model \(f_\theta\) produces a prediction
and, optionally, a rationale or reasoning trace,
$
(\hat y, \hat R) = f_\theta(q,\mathbf{Z}).
$
We call the full input \((q, \mathbf{Z})\) the prompt, and a task is the set of queries that share
one objective and label space, such as disease prediction.

Our objective is to determine whether a biological input \(\mathbf{Z}_i\) is used by the model to make its prediction. We test this by changing the task-relevant biological content of
\(\mathbf{Z}_i\) while holding \(q\) and the remaining inputs \(\mathbf{Z}_{-i}\) fixed, and measuring
prediction sensitivity and changes in task performance. We also test source preference with \emph{evidence conflicts}, inputs in which two
biological evidence sources support different predictions, constructed by combining \(\mathbf{Z}_i\)
from one query with one or more inputs in \(\mathbf{Z}_{-i}\) from another, and measure which source the prediction follows. When perturbing \(\mathbf{Z}_i\) has little effect on task performance, we additionally test whether the target can be predicted from \(\mathbf{Z}_i\) using a
simple predictive baseline, such as a majority-vote lookup or linear probe.
\(\mathbf{Z}_i\) is \emph{underutilized} when it supports prediction of the target with such a baseline, but changing its task-relevant biological content has little effect on
the reasoning model's task performance after controlling for predictive information in
\(\mathbf{Z}_{-i}\) where possible.

\begin{table}[h]
\caption{Models, checkpoints, input sources, tasks, and evaluation data.
Underlined checkpoints are used unless otherwise specified, -- refers only one released checkpoint.
BioReason checkpoints are from~\cite{biopostrain}, as the original study
did not release checkpoints.}
\vspace{-2mm}
\label{tab:models_tasks}
\centering
\scriptsize
\setlength{\tabcolsep}{3pt}
\begin{tabularx}{\linewidth}{
@{}
>{\raggedright\arraybackslash}p{1.7cm}
>{\raggedright\arraybackslash}p{1.3cm}
>{\raggedright\arraybackslash}p{1.5cm}
>{\raggedright\arraybackslash}X
>{\raggedright\arraybackslash}p{1.6cm}
>{\raggedright\arraybackslash}X
@{}
}
\toprule
\textbf{Model} &
\textbf{Modality} &
\textbf{Checkpoints} &
\textbf{Input sources} &
\textbf{Task} &
\textbf{Evaluation data} \\
\midrule

BioReason \citep{fallahpour2025bioreason} checkpoints from \citep{biopostrain}&
Genomic sequence &
SFT, \underline{RL}; 42 post-training checkpoints &
Projected reference and variant Evo2 representations
($\mathbf{Z}_{\mathrm{Evo2}}$);
chromosome ($\mathbf{Z}_{\mathrm{chr}}$),
gene ($\mathbf{Z}_{\mathrm{gene}}$),
KEGG pathway ($\mathbf{Z}_{\mathrm{pathway}}$), and
network ($\mathbf{Z}_{\mathrm{network}}$) text &
Disease prediction &
1,449 KEGG-derived queries over 708 genomes and 37 disease labels;
165 genome-dependent queries; 145 held-out queries for the
auxiliary task evaluation \\
\addlinespace

ChatNT \citep{chatnt} &
Genomic sequence &
-- &
Nucleotide Transformer representations
($\mathbf{Z}_{\mathrm{NT}}$) &
Splice donor, splice acceptor, and TATA promoter prediction &
4,578 ChatNT test queries from three binary tasks (2,000 splice donors, 2,000 splice acceptors, 578 TATA promoters; balanced, up to 1,000 per class) \\
\midrule

BioReason-Pro \citep{bioreasonpro} &
Protein sequence &
SFT, \underline{RL} &
ESM3 ($\mathbf{Z}_{\mathrm{ESM3}}$);
GO-GPT ($\mathbf{Z}_{\mathrm{GO\text{-}GPT}}$);
InterPro ($\mathbf{Z}_{\mathrm{InterPro}}$);
organism text ($\mathbf{Z}_{\mathrm{organism}}$) &
Protein function prediction &
14,102 reviewed human proteins; 91-protein temporal holdout;
540 proteins across 140 InterPro families; 96 GO \texttt{NOT} pairs \\
\addlinespace

Prot2Text-V2 \citep{fei2025prottextv} &
Protein sequence &
-- &
ESM2 ($\mathbf{Z}_{\mathrm{ESM2}}$);
protein name ($\mathbf{Z}_{\mathrm{name}}$);
taxon text ($\mathbf{Z}_{\mathrm{organism}}$) &
Protein function description generation &
3,917 proteins from test split; 632 same-taxon pairs (1,264 evidence conflicts between name and sequence) \\
\midrule

Cell2Sentence-Scale \citep{levine2024cell2sentence} &
Single-cell transcriptome &
2B, \underline{27B} &
Expression-ranked gene sentence
($\mathbf{Z}_{\mathrm{cell}}$) &
Cell type annotation and rationale generation &
Five atlases; 4,846 cells for annotation;
2,000 cells for rationale analysis \\
\addlinespace

CellWhisperer \citep{cellwhisperer} &
Single-cell transcriptome &
-- &
CellWhisperer transcriptome representation
($\mathbf{Z}_{\mathrm{CW}}$);
expression-ranked top-$k$ gene list prefixes
($\mathbf{Z}_{\mathrm{cell}}^{(k)}$), when supplied &
Cell type prediction and evidence conflict &
Same 4,846 C2S-Scale cells (five atlases); 330 same-tissue pairs, with correct prediction from representation only (660 conflicts) \\
\bottomrule
\end{tabularx}
\vspace{-3mm}
\end{table}

\subsection{Evaluation Protocol}
\label{sec:evaluation_protocol}

Table~\ref{tab:models_tasks} summarizes the six models we benchmark, their input sources, model variants, tasks, and evaluation data. We organize our evaluation around four research questions.
\textbf{RQ1: Does task-relevant biological content in a biological input affect the model's prediction?}
We perturb the task-relevant biological content of $\mathbf{Z}_i$ while holding $q$ and
$\mathbf{Z}_{-i}$ fixed and measure changes in predictions and task performance, and we construct evidence
conflicts from unmodified queries to measure source preference. When the perturbation
has little effect on performance, we additionally test whether the target can be predicted from
$\mathbf{Z}_i$ using a simple predictive baseline. When possible, we also remove or control
other predictive biological inputs before concluding that $\mathbf{Z}_i$ is underutilized.
\textbf{RQ2: Does higher benchmark performance correspond to greater performance contributions from biological inputs?}
Across model variants that differ in post-training checkpoint or scale, we compare benchmark
performance with its change under the same biological perturbations used in RQ1. This tests whether
improvements in task performance are accompanied by a greater performance contribution from the evaluated biological input.
\textbf{RQ3: Do reasoning traces and generated explanations reflect the supplied biological evidence?}
We evaluate whether generated responses agree with the biological inputs, and how they change when the biological input is perturbed. For evaluated models that do not produce reasoning traces during prediction, we prompt them to identify the biological evidence supporting their predictions.
\textbf{RQ4: Can auxiliary supervision increase use of a biological input?}
We add an auxiliary objective that requires the model to predict biological content from
$\mathbf{Z}_i$ and test whether this increases the effect of $\mathbf{Z}_i$ on the downstream
task performance.

%% file: iclr2026/031modality-dependence.tex
\vspace{-2mm}
\subsection{RQ1: Foundation model representations can contribute little to performance even when they encode task information}
\label{sec:rq1}\vspace{-2mm}

Here we test prediction sensitivity, performance contribution, and source preference for the evaluated biological inputs with perturbations of biological inputs while holding the query and remaining inputs fixed, and construction of evidence conflicts from unmodified queries. For inputs with limited performance contributions, we confirm their utility to the task with linear probes to test target predictability. %
Appendices~\ref{app:shared} and~\ref{app:rq1} provide the datasets and experimental setup used.

\begin{figure}[b]
    \centering
    \vspace{-6mm}
    \includegraphics[width=\linewidth]{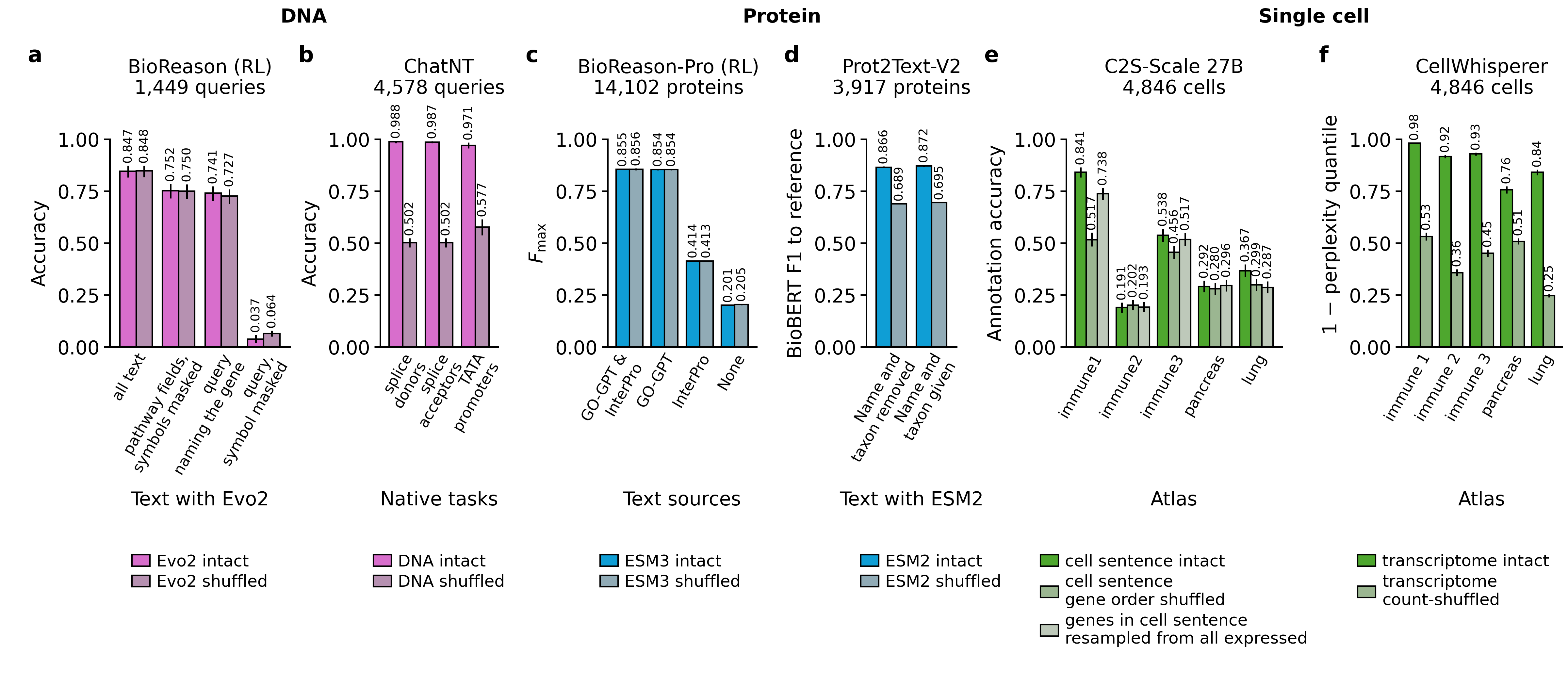}
    \vspace{-10mm}
    \caption{
    \textbf{ Reasoning model performance with representations of intact and shuffled sequences.}
    \textbf{a} BioReason disease prediction accuracy across text conditions.
    \textbf{b} ChatNT accuracy on three native DNA tasks.
    \textbf{c} BioReason-Pro $F_{\max}$
    across text conditions.
    \textbf{d} Prot2Text-V2 BioBERT BERTScore F1 with name and taxon text
    removed or provided.
    \textbf{e} C2S-Scale 27B accuracy with intact, reordered, or resampled
    gene sentences.
    \textbf{f} CellWhisperer $1-\text{perplexity quantile}$ before and after
    permuting counts across genes within each cell.
    Error bars are 95\% CIs.
    }
    \label{fig:rq1_perturbations}
    \vspace{-4mm}
\end{figure}

\paragraph{Biological perturbations reduce performance in four models but have
little effect on performance in BioReason and BioReason-Pro.}
BioReason disease prediction accuracy is almost unchanged with shuffled DNA inputs
(0.847 vs.\ 0.847), although it changes 101 of 1,449 answers (7.0\%, Table~\ref{tab:change-decomposition-dna}). The largest decrease in accuracy occurs when the gene name is removed (Fig.~\ref{fig:rq1_perturbations}a). For ChatNT, accuracy decreases from 0.988 to 0.502 for splice donors, 0.987 to 0.502 for splice acceptors, and 0.971 to 0.577 for TATA promoters (Fig.~\ref{fig:rq1_perturbations}b). The correct label for a shuffled sequence is poorly defined, though this result does support that ChatNT responses differ for the shuffled sequence compared to the original sequence.

For protein models, providing the ESM3 embedding of a shuffled sequence to BioReason-Pro with GO-GPT and InterPro unperturbed results in minimal change in $F_{\max}$ (0.855 vs.\ 0.856) across 14,102 human proteins. On a temporal holdout of proteins annotated after the checkpoint release, this is still observed (0.685 vs.\ 0.682). The shuffled sequence does influence the predicted GO term set for 46.8\% of 14,102 proteins, with a mean Jaccard similarity of 0.89 to the GO term set predicted for the original sequence (Table~\ref{tab:brp-proteins}).
For Prot2Text-V2, BioBERT BERTScore F1 decreases by 0.177 with the name and taxon removed, showing a performance contribution from the ESM2 representation.

\begin{figure}[h]
    \centering
    \vspace{-3mm}
    \begin{minipage}[t]{0.48\linewidth}
        \vspace{-4pt}
        \centering
        \includegraphics[width=\linewidth]
            {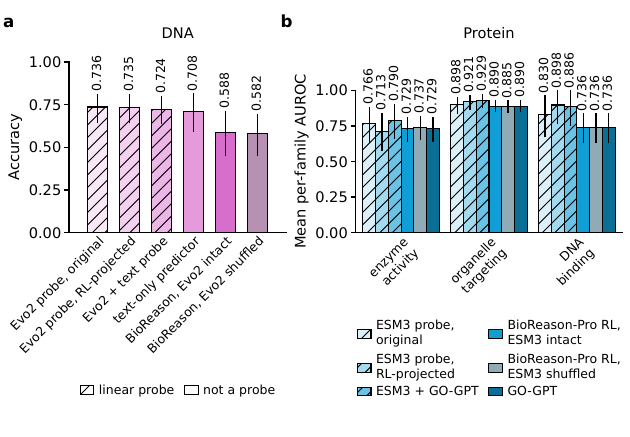}
        \vspace{-8mm}
        \caption{
        \textbf{Targets remain predictable from biological representations.}
        \textbf{a} Disease prediction accuracy on genome-dependent queries
        for linear probes and BioReason.
        \textbf{b} Mean per-family AUROC for linear probes and
        BioReason-Pro. Error bars are 95\% CIs.
        }
        \label{fig:rq1_probes}
    \end{minipage}
    \hfill
    \begin{minipage}[t]{0.48\linewidth}
        \vspace{-3pt}
        \centering
        \includegraphics[width=\linewidth]
            {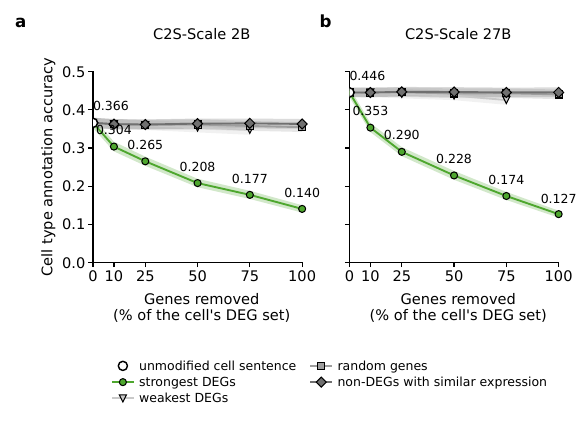}
        \vspace{-8mm}
        \caption{
        \textbf{DEGs contribute to C2S-Scale performance.}
        Accuracy after removing increasing fractions of the
        strongest DEGs compared to other sets of genes for
        2B (\textbf{a}) and 27B (\textbf{b}).
        Bands are 95\% CIs.
        }
        \label{fig:rq1_c2s_degs}
    \end{minipage}
    \vspace{-5mm}
\end{figure}

Across five single-cell atlases, C2S-Scale 27B mean cell type accuracy falls from
0.446 to 0.351 after gene order shuffling and to 0.406 after resampling
genes from all expressed genes (Fig.~\ref{fig:rq1_perturbations}e). In Fig.~\ref{fig:rq1_c2s_degs} we further observe that removing DEGs from the cell sentence reduces C2S-Scale cell type annotation accuracy more than removing non-DEGs with similar expression.
On the same cells we also evaluate CellWhisperer. Permuting counts across gene identities and recomputing CellWhisperer representations reduces $1-\text{perplexity quantile}$ in every atlas, including from 0.98 to 0.53 in immune1 and 0.84 to 0.25 in lung (Fig.~\ref{fig:rq1_perturbations}f). The single-cell inputs contribute to both C2S-Scale and CellWhisperer performance.

\begin{figure}[h]
    \centering
    \vspace{-3mm}
    \includegraphics[width=\linewidth]{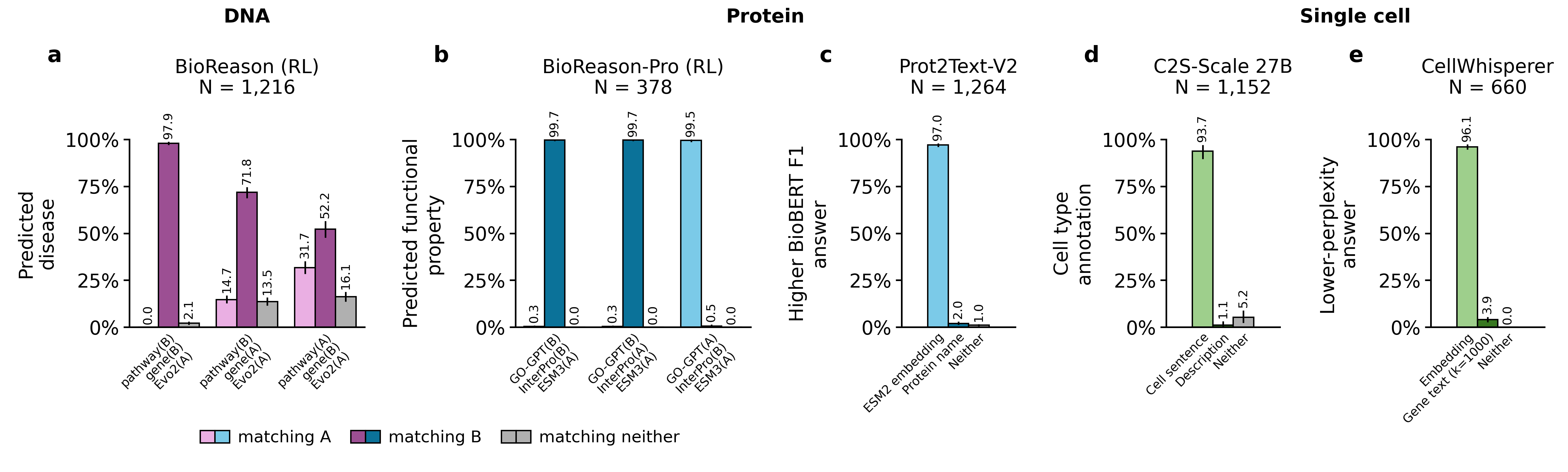}
    \vspace{-8mm}
    \caption{
    \textbf{Evidence conflicts from unmodified queries.}
    \textbf{a} BioReason: $\mathbf{Z}_{\mathrm{Evo2}}$ from $A$,
    gene/pathways from $A$ or $B$.
    \textbf{b} BioReason-Pro: $\mathbf{Z}_{\mathrm{ESM3}}$ from $A$,
    GO-GPT/InterPro from $A$ or $B$.
    \textbf{c} Prot2Text-V2: $\mathbf{Z}_{\mathrm{ESM2}}$ from $A$,
    protein name from B, with taxon fixed.
    \textbf{d} C2S-Scale 27B: $\mathbf{Z}_{\mathrm{cell}}$ from $A$,
    cell type description from $B$.
    \textbf{e} CellWhisperer: $\mathbf{Z}_{\mathrm{CW}}$ from $A$,
    top-1,000 gene text from $B$.
    Bars indicate preference for A, B, or neither. Error bars are 95\% CIs.
    }
    \label{fig:rq1_swap}
    \vspace{-4mm}
\end{figure}

\paragraph{Evidence conflicts show whether models follow text or biological inputs.}
Shuffling input sequences can result in out of distribution inputs, so here we construct evidence conflicts with biological inputs from two different entities combined in the one prompt. For BioReason, Evo2 representations from $A$ and gene and pathway text from $B$ results in disease predictions consistent with $B$ in 97.9\% of conflicts (Fig.~\ref{fig:rq1_swap}a). BioReason-Pro prefers the GO-GPT source, with responses consistent with GO-GPT in 99.7\% of conflicts with InterPro and ESM3 inputs (Fig.~\ref{fig:rq1_swap}b). The preference for text remains in pairs where the protein without the property has an experimental GO \texttt{NOT} annotation, with responses consistent with the GO-GPT and InterPro sources in 98.1\% of 106 conflicts (Fig.~\ref{fig:si_gonot_conflict}). Prot2Text-V2 instead favors the sequence source over the conflicting protein name in 97.0\% of conflicts (Fig.~\ref{fig:rq1_swap}c). C2S-Scale 27B predicts the gene sentence source in 93.7\% of conflicts with another cell type's description (Fig.~\ref{fig:rq1_swap}d). CellWhisperer assigns lower perplexity to the answer supported by the representation in 96.1\% of conflicts with another cell's top-1,000 gene list (Fig.~\ref{fig:rq1_swap}e). ChatNT is not evaluated here because its evaluated queries specify the task without providing query-specific biological text that could conflict with the DNA input. These source preferences are consistent with the performance contributions observed in the perturbation experiments.
\paragraph{Representations with limited performance contributions are predictive of task targets.} We next ask if the foundation model representations that contribute little to BioReason and BioReason-Pro performance are predictive of task targets (Fig.~\ref{fig:rq1_probes}). We fit linear probes on mean-pooled foundation model output and on the same representations after each biological reasoning model's projector.
On 165 disease prediction queries that are genome-dependent (i.e. the text-only input does not always have the same disease label), probes on the Evo2 output and on $\mathbf{Z}_{\mathrm{Evo2}}$ reach 0.736 and 0.735 accuracy, compared with 0.708 for the text-only predictor. Combining Evo2 representations with text features improves the text-only predictor but is below the Evo2-only probes. The probe's improvement is driven by performance on questions with gene PRNP (98 of 165 queries), for which probe accuracy reaches 0.855, compared with 0.708 for the text-only predictor.

For BioReason-Pro, we test whether ESM3 representations separate proteins that share a protein family but differ in function. We select 540 proteins from 140 InterPro families whose members differ in enzyme activity, organelle targeting, or DNA binding, so the family annotation alone does not determine the property. For these properties, the protein without the property lacks the GO annotation. We observe a similar result when we construct pairs where one protein has an annotated GO term and the other protein has the annotated \texttt{NOT} of that GO term (Fig.~\ref{fig:si_gonot_conflict}). Linear probes on the ESM3 representations reach mean per-family AUROCs of 0.766, 0.898, and 0.830 for the three properties, and probes on the projected representations $\mathbf{Z}_{\mathrm{ESM3}}$ reach 0.713, 0.921, and 0.898, compared with 0.729, 0.890, and 0.736 for BioReason-Pro (Fig.~\ref{fig:rq1_probes}b). The joint probe with GO-GPT and ESM3 representations also scores higher than GO-GPT on all three properties.
The GO \texttt{NOT} pairs span 50 GO terms, so they have no label shared across pairs for a probe.
Functional properties are thus predictable from the ESM3 representation.

\begin{tcolorbox}
\textbf{Finding 1. Some biological inputs contribute little to model performance.} Shuffling reveals limited performance contributions, and evidence conflicts reveal preferences for other inputs. However, linear probes show these inputs are predictive of targets for the task.
\end{tcolorbox}

%% file: iclr2026/032model-variants.tex
\subsection{RQ2: Post-training raises overall task accuracy without increasing the biological input's performance contribution}
\label{sec:rq2}
\vspace{-3mm}

We next test whether model variants with higher benchmark performance also show greater performance contributions from
the biological inputs evaluated in RQ1. For BioReason and BioReason-Pro, we compare the released
SFT and RL checkpoints under the same perturbations of $\mathbf{Z}_{\mathrm{Evo2}}$ and
$\mathbf{Z}_{\mathrm{ESM3}}$. For C2S-Scale, we compare the 2B and 27B checkpoints using the
DEG removal experiment in Fig.~\ref{fig:rq1_c2s_degs}.
Appendix~\ref{app:rq2} gives the checkpoints, their training configurations, and the evaluation
settings held fixed across them.

\begin{figure}[h]
    \centering
    \includegraphics[width=\linewidth]{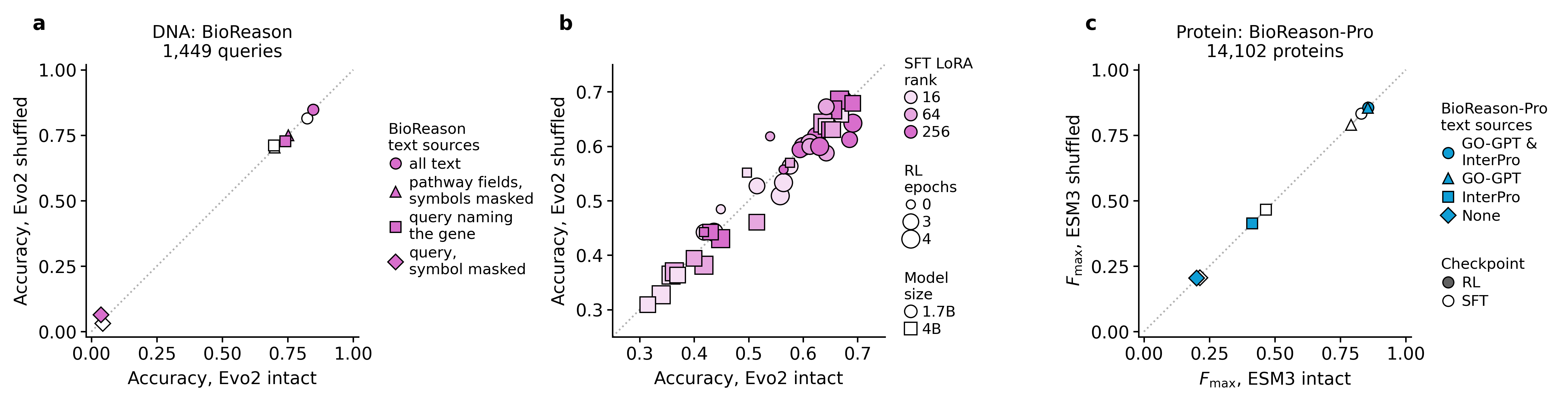}
    \vspace{-4mm}
    \caption{
    \textbf{Biological representations contribute similarly little to SFT and RL checkpoint performance.}
    BioReason accuracy with original or perturbed
    $\mathbf{Z}_{\mathrm{Evo2}}$ across \textbf{a} biological text conditions and
    \textbf{b} for 42 post-training checkpoints on the 165 genome-dependent queries.
    \textbf{c} BioReason-Pro $F_{\max}$ with original or perturbed
    $\mathbf{Z}_{\mathrm{ESM3}}$ across biological text conditions.
    }
    \vspace{-3mm}
    \label{fig:sft_vs_rl}
\end{figure}

\paragraph{SFT and RL improve benchmark performance without greater performance contributions from
$\mathbf{Z}_{\mathrm{Evo2}}$ or $\mathbf{Z}_{\mathrm{ESM3}}$.}
For BioReason, the SFT and RL checkpoints differ in disease prediction accuracy but show similarly limited performance contributions
from $\mathbf{Z}_{\mathrm{Evo2}}$ (Fig.~\ref{fig:sft_vs_rl}a). With all biological text present, SFT
accuracy is 0.823 with the original $\mathbf{Z}_{\mathrm{Evo2}}$ and 0.814 when it is recomputed from shuffled DNA, whereas RL scores 0.847 and 0.847 under the same conditions. The same pattern is observed as we dropout the KEGG pathway and gene name, with both checkpoints remaining close to the
identity line across the four conditions. Thus, the higher disease prediction accuracy of the BioReason RL checkpoint is not accompanied by a greater performance contribution from $\mathbf{Z}_{\mathrm{Evo2}}$.

We observe the same pattern with BioReason-Pro with use of  $\mathbf{Z}_{\mathrm{ESM3}}$ in impacting the $F_\mathrm{max}$ performance (Fig.~\ref{fig:sft_vs_rl}c). When GO-GPT and InterPro both present, RL increases
$F_{\max}$ from 0.829 for SFT to 0.855, while replacing $\mathbf{Z}_{\mathrm{ESM3}}$ with a representation recomputed from shuffled protein sequence changes $F_{\max}$ from 0.829 to 0.833 for SFT and from 0.855 to 0.856 for RL. This limited performance contribution is also observed as InterPro and GO-GPT inputs are removed.
Likewise, RL improves protein function prediction without increasing the performance contribution from $\mathbf{Z}_{\mathrm{ESM3}}$.
The same comparison across 42 post-training checkpoints that vary model size, SFT LoRA rank, RL LoRA rank and RL duration shows accuracy varying without a consistent increase in the advantage of the original $\mathbf{Z}_{\mathrm{Evo2}}$ over the representation recomputed from shuffled DNA (Fig.~\ref{fig:sft_vs_rl}b). Appendix~\ref{app:rq2-sweep} reports the individual sweeps.

\paragraph{DEGs contribute more to performance in the larger C2S-Scale model.}
C2S-Scale provides a contrasting relationship between benchmark performance and biological input contribution. Mean cell type accuracy with the unmodified $\mathbf{Z}_{\mathrm{cell}}$ increases from 0.366 for the 2B checkpoint to 0.446 for 27B while the accuracy decrease under DEG removal from the cell sentence also increases (Fig.~\ref{fig:rq1_c2s_degs}). At 100\% deletion of DEGs, the 2B model scores 0.140 after removing the strongest DEGs, and 0.363 after removing the same number of non-DEGs with similar expression, resulting in a DEG-specific difference of 0.223. The corresponding accuracies for 27B are 0.127 and 0.446, resulting in a larger difference of 0.319. Therefore, the DEG-specific effect is larger for 27B in all five atlases at full DEG removal. Unlike BioReason and BioReason-Pro, the larger C2S-Scale model also shows a greater performance contribution from biological inputs.

\begin{tcolorbox}
\textbf{Finding 2. Improvements in task performance across post-training configurations are not consistently accompanied by a greater performance contribution from the biological input.} RL improves performance for BioReason and BioReason-Pro but does not increase the performance contribution from foundation model representations. Conversely, for cell type annotation C2S-Scale 27B shows a larger accuracy decrease under DEG dropout in the cell sentence than the 2B model.

\end{tcolorbox}

%% file: iclr2026/033rationales.tex
\subsection{RQ3: Reasoning traces and rationales do not reflect biological inputs}
\label{sec:rq3}

We analyze the reasoning traces of BioReason and BioReason-Pro, and prompted rationales from C2S-Scale, to assess correctness, relevance to biological inputs, and consistency with final predictions.

\begin{figure}[h]
    \centering
    \includegraphics[width=\linewidth]{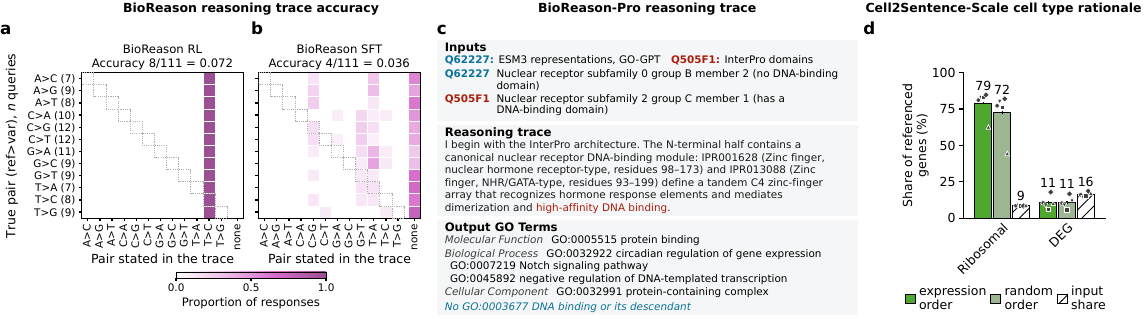}
    \vspace{-4mm}
    \caption{
    \textbf{Analysis of reasoning traces and rationales.}
    The reference-to-variant base pair BioReason states in its trace on the 111 held-out substitution queries for the \textbf{a} RL and \textbf{b} SFT checkpoint.
    \textbf{c} Example BioReason-Pro reasoning trace for an evidence conflict.
    \textbf{d} Proportion of the genes in the C2S-Scale 27B rationale that are ribosomal protein genes and DEGs.
    Dots are the five single-cell atlases and error bars are 95\% CIs.
    }
    \label{fig:rq3_traces}
\end{figure}

\paragraph{BioReason states variant properties that its input does not determine.}
In the BioReason reasoning traces, the first step identifies the substituted bases, a genomic coordinate, the chromosome, and the gene, followed by steps that reason from the gene to a disease. On the 145 held-out queries of the pathway network split, the BioReason models state the correct base for 7.2\% and 3.6\% of all substitution mutations for RL and SFT, respectively (Fig.~\ref{fig:rq3_traces}a,b). Notably, the RL checkpoint states the variant is a substitution and is T>C, on all 145 responses and when the input sequence represented by Evo2 is shuffled. Conversely, the SFT checkpoint is partially responsive to sequence and does not always output the same mutation in its reasoning trace.

\paragraph{BioReason-Pro's final GO predictions omit functional properties stated in its reasoning traces.}
For the evidence conflicts $(A,B)$ with InterPro($B$), GO-GPT($A$), and ESM3($A$) used in RQ1, we test how this affects the reasoning traces when the two members share an InterPro family and have different GO term annotations. We use the InterPro2GO mapping \citep{interpro2go} and filter for evidence conflict pairs where InterPro annotations of $B$ is predictive of the tested functional property (enzyme activity, organelle targeting, DNA-binding) GO term that $A$ is not annotated with (53 out of 189 evidence conflict pairs).
In these responses, the reasoning trace recites all InterPro annotations from $B$. Furthermore in 23 of 53 traces, it states the tested property in words that are not part of a printed InterPro entry name before predicting the function from the GO-GPT input source. For example, the reasoning trace names a DNA-recognition module before emitting a GO set that contains no DNA-binding term (Fig.~\ref{fig:rq3_traces}c).

\paragraph{C2S-Scale rationales show early DEG enrichment but overrepresent ribosomal genes.}
RQ1 shows that removing the strongest DEGs from $\mathbf{Z}_{\mathrm{cell}}$ changes C2S-Scale cell type annotations. We prompt C2S-Scale 27B for annotation rationales on 400 cells from each of five atlases (Appendix~\ref{app:rq3}). Rationales reference 63.4--67.8 genes, of which 93.9--98.7\% occur in $\mathbf{Z}_{\mathrm{cell}}$ and 62.6--84.7\% are ribosomal protein genes (Fig.~\ref{fig:rq3_rationale_supp}a). Mean DEG precision across all referenced genes is 10.63\%, below the random-draw reference, whereas the first five references show enrichment (Fig.~\ref{fig:rq3_rationale_supp}b). Among 452 incorrect cell type annotations naming an atlas type, precision against predicted-type DEGs is 24.2\% versus 9.9\% (Table~\ref{tab:rq3-stratified}), showing that early references can contain markers of the assigned cell type even when the annotation is incorrect. Randomizing gene order reduces the ribosomal share by 6.6 percentage points but remains overrepresented in the rationale and unchanged DEG precision (Fig.~\ref{fig:rq3_traces}d). Marker-focused prompts, ribosomal exclusions, and one-shot examples also yield low overall DEG precision (Appendix~\ref{app:rq3-degprompt}). Removing the strongest DEGs changes the rationale gene set more than removing expression-matched non-DEGs (Fig.~\ref{fig:rq3_rationale_supp}c). Finally, adding a rationale request to otherwise matched annotation prompts lowers pooled accuracy from 0.45 to 0.41 (Table~\ref{tab:rq3-label-change}). Thus, early gene references show marker relevance, but full rationales dilute DEG enrichment and changes annotation behavior.

\begin{tcolorbox}
\textbf{Finding 3. Generated reasoning traces and rationales can misrepresent biological inputs they describe.} Models tend to misstate variants, omit stated functions from predictions. C2S-Scale rationales show early DEG enrichment but overrepresent ribosomal genes overall.
\end{tcolorbox}

%% file: iclr2026/034post-training.tex
\subsection{RQ4: Auxiliary sequence supervision increases sensitivity to foundation model representations with little transfer to disease prediction}
\label{sec:rq4}

\begin{wrapfigure}{r}{0.38\linewidth}
    \centering
    \vspace{-7mm}
    \includegraphics[width=\linewidth]{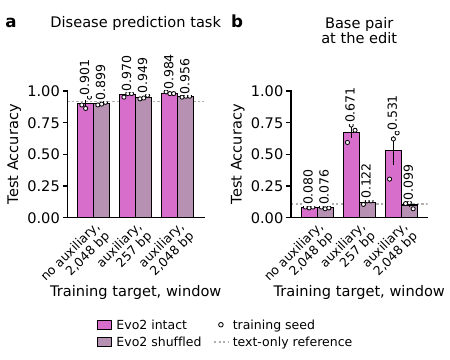}
    \vspace{-6mm}
    \caption{
    \textbf{Auxiliary sequence supervision for BioReason.}
    \textbf{a} Disease prediction accuracy and \textbf{b} edited base pair accuracy on 145 held-out pathway-network split
    queries with $\mathbf{Z}_{\mathrm{Evo2}}$ derived from original or shuffled DNA.
    Error bars the standard error over three training seeds.
    }
    \vspace{-3mm}
    \label{fig:rq4_auxiliary}
\end{wrapfigure}

BioReason is trained with supervised fine-tuning on reasoning traces that specify the base change in the first step, yet RQ3 shows that it rarely generates the correct base change. We therefore test whether explicit supervision of the edited bases improves their prediction and increases the contribution of $\mathbf{Z}_{\mathrm{Evo2}}$ to disease prediction performance. We replace the full reasoning trace with a short target specifying the reference and variant bases at the edited position while retaining the disease answer (Appendix \ref{app:rq4}). On 145 held-out queries, auxiliary supervision with 257-bp input windows yields edited base pair accuracy of 0.671 with intact DNA and 0.122 after shuffling. With 2,048-bp windows, the corresponding accuracies are 0.531 and 0.099, compared with 0.080 for the base changes stated without auxiliary supervision (Fig. \ref{fig:rq4_auxiliary}b). Both auxiliary models also respond to sequence replacement in evidence conflicts (Fig. \ref{fig:rq4_auxiliary_matrices}), while disease prediction accuracy changes little under shuffling (Fig. \ref{fig:rq4_auxiliary}a). However, 135 of the 145 held-out queries contain genomes present in training. Pooling validation and held-out queries, edited base pair accuracy on 218 queries with previously seen genomes is 0.691 and 0.606 for the 257-bp and 2,048-bp models, respectively. On 72 queries with unseen genomes, accuracy falls to 0.148 and 0.106, compared with 0.111 for always predicting the most frequent training base pair (Table \ref{tab:rq4-unseen}). Thus, the improvement is concentrated in previously seen genomes and increased sequence sensitivity on the original split does not establish reliable edited-base prediction on unseen genomes.

\begin{tcolorbox}
\textbf{Finding 4. Auxiliary supervision increases BioReason's sequence sensitivity, but edited base prediction generalizes poorly to unseen genomes.}
Edited-base predictions respond to DNA shuffling and genome replacement, but accuracy improvements are concentrated to genomes present during training. Disease prediction accuracy also changes little when Evo2 representations are recomputed from shuffled DNA.
\end{tcolorbox}

%% file: iclr2026/040discussion.tex
Biological reasoning benchmarks can reward accurate answers without requiring models to interpret the biological representations they receive. When accompanying annotations already support the target, success does not establish that a model can extract complementary information from sequence or cellular inputs. Sensitivity to those inputs can occur but have limited overall performance benefit. Higher accuracy after post-training does not indicate a greater performance contribution from the biological input. Generated reasoning traces may also not accurately reflect inputs to the reasoning model.

\xhdr{Limitations and future work}
We evaluate six biological reasoning models on specific tasks and biological inputs, so the findings may not generalize to other architectures or biological modalities. Some analyses shuffle the sequence, which can move representations outside the distribution of natural sequences, although evidence conflicts built from unmodified inputs support the same conclusions. Linear probes from representations given to the reasoning model show that a target is predictable, but they do not show how the language model can use that representation. Future work could evaluate additional models and test training prompt variants within the same model while holding other training settings fixed. Reinforcement learning objectives that reward correct responses to changes in biological inputs, in addition to final accuracy, are another direction to evaluate.

\xhdr{Conclusion}
Biological reasoning models can achieve high benchmark performance with limited contributions from
predictive biological inputs. Our results show that neither access to a predictive representation nor
improved performance after SFT or RL ensures a performance contribution from that
representation. The design and evaluation of biological reasoning models should consider the correct use of task-relevant biological inputs as an additional objective alongside predictive accuracy.

%% file: iclr2026/091appendix_rq1.tex
\section{Datasets, Models, and Evaluation Methods}\label{app:shared}

This appendix section records the datasets, checkpoints, prompts, scoring rules and uncertainty estimates used.

\subsection{Evaluation Datasets and Preprocessing}\label{app:datasets}

\subsubsection{Genomic datasets and evaluation splits}\label{app:data-dna}

\paragraph{BioReason disease prediction task.}
The task uses the KEGG-derived dataset released with BioReason
\citep{fallahpour2025bioreason} aggregated over all three splits (1,159 train, 146 test and 144 validation queries). The 1,449 queries cover 708 distinct reference/variant genome pairs, 186 distinct prompt texts and 37 disease labels. Each query comes with its prompt text, a reference sequence, a
variant sequence and one gold disease. The prompt text has four variable fields: a
\texttt{Chromosome Number}, a \texttt{Network Definition of the pathway} in which the
variant gene is marked with an asterisk, \texttt{Genes in the pathway} with \texttt{SYMBOL; description} entries, and a query that names the variant gene and asks
which disease the allele contributes to. The pathway definitions, pathway genes and disease labels are taken from KEGG. The dataset sequences for the reference and variant are encoded with Evo2. The disease label is the dataset's answer string, and the 37 labels form the closed vocabulary used for scoring (Appendix~\ref{app:models}). All 1,449 queries are evaluated because every comparison in Figure~\ref{fig:rq1_perturbations}a is paired with the same query but with the input shuffled.

\paragraph{ChatNT tasks.}
ChatNT is evaluated on three of its own binary tasks from the released test split which contains three tasks: (1) splice donors (2,198 test queries, 1,098 Yes and 1,100 No, 600\,bp), (2) splice acceptors (2,218 queries, 1,118 and
1,100, 600\,bp) and (3) TATA promoters (621 queries, 332 and 289, 300\,bp). From each task we sample up to
1,000 queries per class, giving 2,000, 2,000 and 578 queries (4,578 in total, balanced within task).

\subsubsection{Protein datasets and annotation processing}\label{app:data-protein}

\paragraph{Reviewed human proteome (BioReason-Pro).}
Proteins are obtained from UniProt with organism ID 9606 and filtered for reviewed entries. We keep only entries with curated GO ground truth, sequences of 120 to 1,024 residues (1,024 amino acids is the maximum sequence length for the GO-GPT encoder). This leaves 14,102 proteins, of
which 14,019 have at least one InterPro entry. Ground truth is the GOA human annotation file (GAF 2.2, generated 2026-07-28), restricted to the evidence codes EXP, IDA, IPI, IMP, IGI, IEP, TAS and IC, with NOT qualifiers dropped and high-throughput codes excluded, over all three
aspects. The file holds 16,991 annotated proteins with 149,820 annotations (8.8 per protein).

\paragraph{Temporal holdout (BioReason-Pro).}
We use a cutoff of 2026-03-20, given by the day the BioReason-Pro preprint was posted. Proteins included in this holdout are reviewed UniProt entries of 120 to
1,024 residues added on or after the cutoff (706 entries). Entries whose GO cross-references carry only electronic evidence (615) are dropped, leaving 91 proteins, 90 of them
with an InterPro entry. Labels are the UniProt GO cross-references with non-electronic evidence. The set spans several organisms (22 \emph{Drosophila}, 18 \emph{Taxus chinensis}, 9 \emph{Danio} among others), and its proteins carry a median of 2 curated terms against 8.8 on the
proteome. The level of $F_{\max}$ on the holdout is therefore not comparable with the proteome, and
we interpret only the within-holdout contrast between intact and shuffled input.

\paragraph{Prot2Text-V2 test split.}
Prot2Text-V2 is evaluated on the official test split containing 4,203 proteins. The released pipeline truncates sequences to 1,021 residues. We exclude the
286 proteins longer than 1,021 residues so that a shuffled sequence is a permutation of the whole
input, leaving 3,917 proteins for the perturbation experiment.

\subsubsection{Single-cell atlases and cell selection}\label{app:data-sc}

Cell sentences are generated by us from the five atlases that the C2S-Scale authors use for cell type
annotation (Table~\ref{tab:sc-atlases}). Each atlas is prepared as follows: the matrix holds
$\log(1+x)$ of the raw counts, gene names are upper-cased HGNC symbols, and the annotation is the atlas's own
cell type column. Cells which have at least {1,000} expressed genes are drawn as a balanced sample across annotated types excluding cells annotated \texttt{unknown} from each atlas. This results in 900, 1,049, 957, 939, and 1,001 cells for immune1, immune2, immune3, pancreas, and lung, and 4,846 cells in total.
The 2,000 cells used for rationale analysis (Table~\ref{tab:models_tasks}) are 400
cells per atlas drawn without replacement from the 4,846 cells. CellWhisperer is evaluated on the same 4,846 cells.

\begin{table}[h]
\caption{The five single-cell atlases. Shorthand labels are study names such as \texttt{alpha}, \texttt{beta},
\texttt{Basal} and \texttt{Ciliated}.}
\label{tab:sc-atlases}
\centering\scriptsize
\setlength{\tabcolsep}{3pt}
\begin{tabular}{llrrrl}
\toprule
Atlas & Accession & Cells & Types & Options & Label source \\
\midrule
immune1  & CELLxGENE \texttt{62ef75e4}; E-MTAB-11536 & 30,423 & 35 & 34 & Cell Ontology names \\
immune2  & CELLxGENE \texttt{e9360edf}; GSE271896    & 33,923 & 38 & 38 & Cell Ontology names \\
immune3  & CELLxGENE \texttt{cc431242}; GSE299043    & 30,593 & 31 & 31 & Cell Ontology names \\
pancreas & GSE84133 (Baron et al.\ 2016)             &  5,641 & 14 & 14 & study shorthand \\
lung     & GSE280502 (Kim et al.\ 2024)              &  6,160 &  7 &  7 & study shorthand \\
\bottomrule
\end{tabular}
\end{table}

For C2S-Scale we ask the model to select from an option list of all cell types in the atlas of the evaluated cell sentence. CellWhisperer is scored by the likelihood of a candidate answer string using the Tabula Sapiens format.

\subsection{Evaluated models}\label{app:models}

\paragraph{BioReason.}
The RL checkpoint is the \path{bio-posttrain-qwen3-1.7b-dna-rl} checkpoint \citep{biopostrain}, a
Qwen3-1.7B model with the GRPO LoRA adapter (rank 16, $\alpha=32$, 1,156 steps) merged, paired with
a frozen Evo2 \texttt{evo2\_1b\_base} encoder \citep{evo2}. The encoder uses a trained projection to the 2,048-d text width. Reference and variant bases are encoded separately, projected, included into $2 \times 2{,}048$ positions of the prompt. $\mathbf{Z}_{\mathrm{Evo2}}$
denotes the projected rows. The linear probes of Appendix~\ref{app:rq1-probe-dna} read the encoder
output at layer \texttt{blocks.20.mlp.l3} (1,920-d) and $\mathbf{Z}_{\mathrm{Evo2}}$, in each case
mean-pooled over the 2,048 positions of the reference and of the variant block and concatenated. We also evaluate the SFT checkpoint, \texttt{mims-harvard/bio-posttrain-qwen3-1.7b-dna-sft} for RQ2. The input to BioReason is the two DNA representations of the wild type and variant followed by the same prompt text format, as in
the released training code.

For disease prediction, the generated disease is lower-cased, stripped of punctuation other than hyphens, apostrophes and slashes, and mapped onto the 37-label vocabulary by substring containment with the
longest label tried first. If the answer segment maps to nothing the whole generation is searched.
A generation that maps to no label is wrong (1.5\% of wild-type queries for the RL checkpoint).
Accuracy is reported as the mean over all queries.

\paragraph{ChatNT.}
We use the released checkpoint which contains a 500M-parameter Nucleotide Transformer v2 encoder that tokenises the sequence as 6-mers (101 tokens for 600\,bp, 51 for 300\,bp), a Perceiver resampler produces 64 representations that are inserted at the DNA placeholder of the
English prompt, and a frozen Vicuna-7B decoder produces the answer. $\mathbf{Z}_{\mathrm{NT}}$
denotes the 64 inserted representations.

The prompt is the released pipeline's context (``A chat between a curious user and an artificial
intelligence assistant that can handle bio sequences \ldots USER: ''), the query
with its DNA placeholder (for example ``Is a donor splice site present in the sequence
\texttt{<DNA>}?''), and ``ASSISTANT:''. The score is the log-softmax over the first answer token
at the last prompt position, with the predicted label being the larger value over the answer tokens `` Yes'' and `` No''.

\paragraph{BioReason-Pro.}
We use the released RL checkpoint \citep{bioreasonpro}. ESM3 (\texttt{esm3\_sm\_open\_v1}) is run
inside generation on the sequence string its representations are projected and spliced at the protein placeholder, and the GO-GPT text is derived from a separate
sequence field. The projector is a two-layer perceptron with a GELU nonlinearity that maps the
layer-37 states ($(L+2) \times 1{,}536$) to the 2,560-d text width, and
$\mathbf{Z}_{\mathrm{ESM3}}$ denotes its output. The probes of Appendix~\ref{app:rq1-probe-protein}
read the layer-37 states and $\mathbf{Z}_{\mathrm{ESM3}}$, each mean-pooled over all $L+2$ positions. We also evaluate the SFT checkpoint in RQ2.

The prompt is BioReason-Pro's own CAFA5 reasoning template with context: ``Given the protein above
from organism \{organism\} with the  following InterPro annotations: \{interpro\} And the following
initial GO term speculations: \{go\_speculations\} Reason about the function of the protein and
focus more on its Molecular Function, Biological Process, Cellular Component. Summarize in UniProt
format.'' The organism string is UniProt's organism name. The prediction is the set of GO identifiers outputted by the model.

Predicted GO sets are scored with the released \texttt{cafaeval} engine in the configuration BioReason-Pro uses (CAFA normalisation, maximum propagation of predictions and ground truth to the
ontology roots over \texttt{is\_a} and \texttt{part\_of}, per-aspect scores averaged over the
aspects present). We report \texttt{cafaeval}'s IA-weighted $F_{\max}$.

For the evidence conflicts each protein carries a specification over one GO term (required or
forbidden). A prediction satisfies a specification when its propagated GO set contains a descendant
of every required term and no descendant of a forbidden term, evaluated as a search over the
ontology graph rather than as set membership.

\paragraph{Prot2Text-V2.}
The checkpoint is \texttt{Prot2Text-V2-11B-Instruct-hf}
\citep{fei2025prottextv}, with the ESM2 encoder \texttt{esm2\_t36\_3B\_UR50D} and a Llama-3.1-8B
decoder.

The prompt is the released system message (``You are a scientific assistant specialized in protein
function predictions \ldots'') and the user message ``Protein name: \{name\} ; Taxon: \{taxon\} ;
Sequence representations: '' followed by the ESM2 tokens.

Each generated description is scored against the protein's UniProt function text with BERTScore F1
using \texttt{dmis-lab/biobert-large-cased-v1.1} at layer 24, after retokenising both texts to
495 tokens, which is the released benchmark script's setting. Scores for shuffled sequences are the mean
over three shuffle seeds. For evidence conflicts, the same score is computed against both reference
functions (Appendix~\ref{app:rq1-conflict}).

\paragraph{Cell2Sentence-Scale.}
The checkpoints are \texttt{C2S-Scale-Gemma-2-27B} and
\texttt{-2B} \citep{levine2024cell2sentence}. The biological input is the cell sentence
$\mathbf{Z}_{\mathrm{cell}}$, the cell's gene symbols in descending expression order, zero-count
genes omitted, truncated at 1,000 genes.

The multiple-choice prompt is built from the opening sentence and answer cue of C2S-Scale's own
cell type annotation template:
\begingroup\scriptsize
\begin{verbatim}
The following is a list of {num_genes} gene names ordered by descending expression
level in a Homo sapiens cell. Your task is to give the cell type which this cell
belongs to based on its gene expression.
Choose your answer from the following list of cell types:
- {option 1}
- {option 2}
...
Cell sentence: {cell_sentence}.
The cell type corresponding to these genes, copied verbatim from the list above, is:
\end{verbatim}
\endgroup
\noindent
The option list is the atlas's full label vocabulary, shuffled for each cell sentence. Options precede the cell sentence.
In evidence conflicts a \texttt{Context:} line holding the description of the cell type is inserted
between the task sentence and the option list. The
generated cell type name is mapped to an option by exact match, then case- and punctuation-normalised match,
then unique containment, then the uniquely longest contained option, then prefix. A generation
that names no offered option is counted as non-compliant and wrong.

\paragraph{CellWhisperer.}
The transcriptome encoder is the released CLIP checkpoint \texttt{cellwhisperer\_clip\_v1.ckpt}
\citep{cellwhisperer}, with the Geneformer model as the transcriptome encoder. Raw counts are
normalised per cell to 10,000, divided by the released gene medians, ranked, and truncated at
2,048 tokens by the released processor. The 2,048-d transcriptome projection is L2-normalised, as
in the released training pipeline, and denoted $\mathbf{Z}_{\mathrm{CW}}$. The chat model is the
released default checkpoint (\texttt{results/chat\_models/default}), a LLaVA model on
Mistral-7B-Instruct-v0.2 whose two-layer projector maps $\mathbf{Z}_{\mathrm{CW}}$ to 8 prompt
tokens.

Prompts use the released Mistral-instruct conversation
template. For queries with input gene text, the prompt is prefixed by the top-genes exchange. The answer tokens of the final assistant turn are scored by the mean log-likelihood per token. The lowest perplexity answer is taken as the model's choice.

The released CellWhisperer evaluation ranks the perplexity of the correct answer under the cell's
own transcriptome against its perplexity under 30 randomly drawn transcriptomes of other cell
types. For cell $i$ with label $L_i$ and
transcriptome condition $c$ (intact or count-shuffled), the score is the share of cells $j$ in the
same atlas with $L_j \neq L_i$ for which the per-token log-likelihood of the answer ``This cell is
a $L_i$'' under $\mathbf{Z}_i^{(c)}$ exceeds its log-likelihood under the intact
$\mathbf{Z}_j$. We report the 1 minus perplexity quantile. The candidate answers are the atlas's canonical labels (Appendix~\ref{app:data-sc}). Evidence conflicts are scored with the same per-token quantity to compare the two candidate answers of a pair.

\xhdr{Implementation}\label{app:repro}
All checkpoints were run on NVIDIA H100 GPUs. We evaluate released checkpoints on released data and change no model weights except in the
auxiliary supervision experiment of Section~\ref{sec:rq4}. Appendix~\ref{app:shared} documents the evaluation datasets and their preprocessing, prompts used,  and evaluation methods, and Appendices~\ref{app:rq1}
to~\ref{app:rq4} document the additional analyses for each RQ. The evaluated weights are available on the HuggingFace Hub from
\path{mims-harvard/bio-posttrain-qwen3-1.7b-dna-sft} and \path{-rl} for BioReason, \path{wanglab/bioreason-pro-sft} and \path{wanglab/bioreason-pro-rl} for
BioReason-Pro, \path{InstaDeepAI/ChatNT}, \path{xiao-fei/Prot2Text-V2-11B-Instruct-hf},
\path{vandijklab/C2S-Scale-Gemma-2-2B} and \path{-27B}, and the CellWhisperer CLIP and chat
checkpoints which are publicly distributed by its authors at \url{https://medical-epigenomics.org/papers/schaefer2025cellwhisperer}. They read the
biological encoders \path{arcinstitute/evo2_1b_base}, \path{EvolutionaryScale/esm3-sm-open-v1},
Nucleotide Transformer v2 500M and Geneformer 12L-30M. We run each model through its own released code:
\url{https://github.com/bowang-lab/BioReason} with the post-training code of
\url{https://github.com/mims-harvard/bio-posttrain},
\url{https://github.com/bowang-lab/BioReason-Pro},
\url{https://github.com/instadeepai/nucleotide-transformer},
\url{https://github.com/ColinFX/Prot2Text-V2},
\url{https://github.com/vandijklab/cell2sentence} and
\url{https://github.com/epigen/cellwhisperer}. The evaluation data is the disease prediction task of \path{wanglab/kegg} \citep{fallahpour2025bioreason}, with the pathway-network split of
\citet{biopostrain} for Section~\ref{sec:rq4}, the ChatNT test tasks of
\path{InstaDeepAI/ChatNT_training_data}, reviewed human proteins from UniProt with their InterPro
and GO annotations, five single cell atlases from CELLxGENE and GEO accessions \path{GSE84133} and
\path{GSE280502}, and the full Cell Ontology release for the cell type descriptions.
The code for recreating all results and figures in this paper is available at \href{https://github.com/mims-harvard/bio-mirage}
{\textcolor{magenta}{\nolinkurl{https://github.com/mims-harvard/bio-mirage}}}.

\section{RQ1: Biological Input Perturbations, Evidence Conflicts, and Predictive Baselines}\label{app:rq1}

\subsection{Biological Input Perturbations}\label{app:rq1-perturb}

\paragraph{DNA.}
The DNA sequence is permuted, so length and base composition are preserved but the edited position is not. Both the reference and the variant
block are shuffled. The shuffled sequences are encoded and projected into the prompt. The prompt text is kept the same.

\paragraph{Protein.}
For BioReason-Pro, the amino acid sequence is permuted then embedded by ESM3 while for GO-GPT, InterPro, and organism we use the wild type sequence. For Prot2Text-V2 the string is shuffled and encoded by ESM2.

\begin{table}[h]
\caption{\textbf{BioReason prediction changes under DNA shuffling.} Each of the 1,449 disease prediction queries falls in exactly one cell, by whether the label the scorer maps the generation to is correct with the input intact and after both Evo2 windows are shuffled. Counts, with per cent of 1,449 in brackets. Net is the arm's accuracy minus the intact accuracy, which equals (incorrect$\rightarrow$correct $-$ correct$\rightarrow$incorrect)$/1{,}449$. Re-decoding the identical prompt changes 0 of 1,449 labels at both checkpoints.}
\label{tab:change-decomposition-dna}
\centering
\small
\setlength{\tabcolsep}{6pt}
\begin{tabular}{lcc}
\toprule
\textbf{Label transition, intact $\rightarrow$ shuffled} & \textbf{RL} & \textbf{SFT} \\
\midrule
correct $\rightarrow$ correct & $1{,}192$ (82.3) & $1{,}094$ (75.5) \\
correct $\rightarrow$ incorrect & $35$ (2.4) & $99$ (6.8) \\
incorrect $\rightarrow$ correct & $36$ (2.5) & $86$ (5.9) \\
incorrect $\rightarrow$ incorrect, same label & $156$ (10.8) & $126$ (8.7) \\
incorrect $\rightarrow$ incorrect, other label & $30$ (2.1) & $44$ (3.0) \\
\midrule
Accuracy, intact & 0.847 & 0.823 \\
Accuracy, shuffled & 0.847 & 0.814 \\
\textbf{Net accuracy change} & $\mathbf{+0.001}$ & $\mathbf{-0.009}$ \\
\bottomrule
\end{tabular}
\end{table}

\begin{table}[h]
\caption{\textbf{Proteins whose BioReason-Pro output changes under ESM3 shuffling.} The GO terms emitted from the intact input are compared with those emitted from the shuffled input, with the GO-GPT and InterPro text retained in both. The first row and the Jaccard are over all 14,102 proteins; the $F_1$ rows are over the 14,086 with a scorable aspect, where $F_1$ is the per-protein information-accretion-weighted $F_1$ against that protein's GOA annotation set propagated over the ontology.}
\label{tab:brp-proteins}
\centering
\small
\setlength{\tabcolsep}{8pt}
\begin{tabular}{lr}
\toprule
\textbf{Quantity} & \textbf{Value} \\
\midrule
Proteins whose emitted GO set changes & $6{,}593$ (46.8\%) \\
Proteins whose $F_1$ improves & $1{,}575$ (11.2\%) \\
Proteins whose $F_1$ worsens & $1{,}527$ (10.8\%) \\
Proteins whose $F_1$ is unchanged & $10{,}984$ (78.0\%) \\
Mean Jaccard of the emitted GO term sets & $0.889$ \\
\bottomrule
\end{tabular}
\end{table}

\paragraph{Gene sentence and transcriptome perturbations.}\label{app:rq1-cell}

For C2S-Scale, gene order shuffling permutes the positions of the cell's top 1,000 genes. Resampling from all expressed genes draws 1,000 genes uniformly without replacement from all of the cell's expressed genes and lists them in the cell's own expression order, so the length and the rank structure are preserved and the identity of the visible genes changes. The resampled window retains a mean of 0.57 to 0.59 of the intact top-1,000 genes in the immune atlases and 0.14 in lung, where a cell expresses a median of 7,709 genes. The cell sentence is used with the same query and scored against the annotated cell type of the original cell sentence.

For CellWhipserer, we permute the cell's full count vector across genes. The permuted counts pass through the Geneformer processor (normalisation, median scaling, ranking, truncation at 2,048 tokens) and CLIP tower, so $\mathbf{Z}_{\mathrm{CW}}$ is recomputed from the permuted gene counts. The
prompt is the representation-only prompt of Appendix~\ref{app:models}, and the scored quantity is the perplexity quantile of Appendix~\ref{app:models} against the intact representations of the other cells of the same atlas.

\paragraph{Biological text removal and masking.}\label{app:rq1-text}

Table~\ref{tab:text-conditions} provides for each text condition drawn in
Figures~\ref{fig:rq1_perturbations} and \ref{fig:rq1_swap}.

\begin{table}[h]
\caption{Text fields present in each text condition. \checkmark\ present, $\times$ removed,
M masked to \texttt{GENE}.}
\label{tab:text-conditions}
\centering\scriptsize
\setlength{\tabcolsep}{3pt}
\begin{tabular}{llcccc}
\toprule
Model & Condition (figure label) & \multicolumn{4}{c}{Fields} \\
\midrule
BioReason & & Chromosome & Network & Gene list & Gene in query \\
\cmidrule(lr){3-6}
& all text (\texttt{wt}) & \checkmark & \checkmark & \checkmark & \checkmark \\
& pathway fields, symbols masked (\texttt{no\_gene}) & \checkmark & M & M (descriptions kept) & M \\
& query naming the gene (\texttt{no\_pathway}) & $\times$ & $\times$ & $\times$ & \checkmark \\
& query, gene symbol masked (\texttt{no\_textkey}) & $\times$ & $\times$ & $\times$ & M \\
\midrule
BioReason-Pro & & GO-GPT & InterPro & Organism & \\
\cmidrule(lr){3-5}
& GO-GPT \& InterPro (\texttt{wt}) & \checkmark & \checkmark & \checkmark & \\
& GO-GPT (\texttt{gogpt\_only}) & \checkmark & $\times$ & \checkmark & \\
& InterPro (\texttt{interpro\_only}) & $\times$ & \checkmark & \checkmark & \\
& None (\texttt{no\_symbolic}, \texttt{no\_modality}) & $\times$ & $\times$ & \checkmark & \\
\midrule
Prot2Text-V2 & & Protein name & Taxon & & \\
\cmidrule(lr){3-4}
& name and taxon given (\texttt{native}) & \checkmark & \checkmark & & \\
& name and taxon removed (\texttt{seq\_only}) & $\times$ & $\times$ & & \\
\midrule
CellWhisperer & & Top-1,000 gene text list & & & \\
\cmidrule(lr){3-3}
& representation only (Fig.~\ref{fig:rq1_perturbations}f) & $\times$ & & & \\
& representation and gene text (Fig.~\ref{fig:rq1_swap}e) & \checkmark & & & \\
\bottomrule
\end{tabular}
\end{table}

\subsection{Evidence Conflicts}\label{app:rq1-conflict}

\paragraph{BioReason.}
We only use queries the model answers correctly with the unmodified input. Three conflicts from $(A,B)$ are constructed, all with both DNA representations from A: pathway input and the gene named in the query from B (1,216 conflicts), pathway input from B with the gene named in the query from A (1,185), and pathway input from A with the gene named in the query from B (1,179). The counts differ because a query is dropped when its answer names no disease in the label set, on 11, 40 and 48 queries. Fig.~\ref{fig:rq1_swap}a is labelled with the largest of the three. The pathway input is the chromosome, network and gene list inputs. It matches A when it equals A's disease label, matches B when it equals B's disease label, and is neither otherwise.

\begin{figure}[h]
    \centering
    \includegraphics[width=0.36\linewidth]{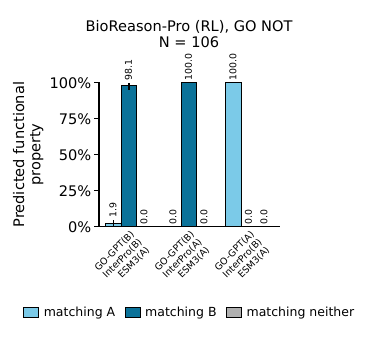}
    \caption{
    \textbf{Evidence conflicts for BioReason-Pro (RL) on GO \texttt{NOT} pairs.}
    $\mathbf{Z}_{\mathrm{ESM3}}$ from $A$, GO-GPT/InterPro from $A$ or $B$, where one protein of each pair has an experimental annotation to a GO term and the other an experimental \texttt{NOT} annotation to the same term. 106 conflicts from 53 pairs. Bars indicate preference for $A$, $B$, or neither. Error bars are 95\% CIs.
    }
    \label{fig:si_gonot_conflict}
\end{figure}

\paragraph{BioReason-Pro.}
A pair is two wild-type proteins of the same organism that share an InterPro family entry and differ in one functional property, scored by one readout GO term. We use four properties: enzyme vs. pseudoenzyme (111 pairs), organelle targeting with a N-terminal transit peptide or not (122 pairs), DNA binding vs. not-binding (76 pairs, \texttt{GO:0003677}) and GO \texttt{NOT} (96 pairs). For the first three properties, one protein carries the readout term in its curated UniProt GO annotation and the other lacks it, so the negative member is defined by the absence of the annotation. The pseudoenzyme is additionally named ``Inactive'' in UniProt or listed as a non-peptidase homologue in MEROPS. These three properties contain 540 proteins. For GO \texttt{NOT} pairs, drawn from the human GOA file, one protein has an experimental annotation to a GO term and the other has an experimental \texttt{NOT} annotation to the same term (evidence codes EXP, IDA, IPI, IMP, IGI or IEP) and no UniProt annotation to the term or its descendants. These 96 pairs contain 150 proteins and span 50 GO terms, so no probe can be  meaningfully trained on them (Appendix~\ref{app:rq1-probe-protein}). A pair is used for the the evidence conflict experiment only if the model correctly annotates both proteins in the pair resulting in 58, 94 and 37 pairs and each contributes two directional conflicts, giving 378 evidence conflicts. For GO \texttt{NOT}, 53 pairs pass and give 106 evidence conflicts (Fig.~\ref{fig:si_gonot_conflict}). For the pair $(A,B)$ with $\mathbf{Z}_{\mathrm{ESM3}}$ from A, three conflicts are constructed: GO-GPT and InterPro both from B, InterPro from B with GO-GPT from A, and GO-GPT from B with InterPro from A. The organism string follows A and is the same for both members. A prediction matches A when it satisfies A's specification and not B's, matches B when it satisfies B's and not A's.

\paragraph{Prot2Text-V2.}
We use unordered pairs of the 3,900 test proteins that share a taxon, correctly annotates both proteins in the pair, has different protein names, each protein name scores higher against its own function than against the partner's, and for both members the annotation from the complete prompt scores higher against its own function than against the partner's. This results in 1,264 directional conflicts. We run the following evidence conflicts: $\mathbf{Z}_{\mathrm{ESM2}}$ of A with B's name, and $\mathbf{Z}_{\mathrm{ESM2}}$ of B with A's name, taxon unchanged. The generated description is scored by BioBERT BERTScore F1 against A's and against B's reference function. It matches the sequence source when its score against the sequence protein's function exceeds its score against the name protein's function by more than 0.005, matches the name source in the opposite case, and is neither when the two scores lie within 0.005 of each other.

\paragraph{C2S-Scale.}
Pairs are obtained from the immune3 atlas as it has the largest number of valid pairs. Cell type descriptions are obtained from Cell Ontology relationship axioms, covering surface markers, capabilities, anatomical location and developmental origin with a maximum of six facts per description. Ancestor and descendant pairs are excluded, and a pair needs at least two contrastive and three shared facts. A pair is valid if the description of B alone makes the 27B model select B over A in a two-way choice.
Conflicts are scored on the 1,152 cells the model annotates as A from the cell sentence alone.
The cell sentence of a type-A cell is combined with a description of type B and all cell type labels from the atlas are given. The predicted cell type matches A when it equals the cell's annotated label, matches B when it equals the partner type, and is neither otherwise.

\paragraph{CellWhisperer.}
In each atlas 200 pairs of cells with different labels are formed by preferring pairs from the same tissue and donor, then the same tissue, then the same donor, and from the same tissue. We keep pairs for which each member is identified correctly from its complete input, $\mathbf{Z}_{\mathrm{CW}}$ together with its own top-1,000 gene list. This holds for 330 pairs, giving 660 directional conflicts. We $\mathbf{Z}_{\mathrm{CW}}$ of A with $k = 1{,}000$ from B's gene list and vice versa. For each conflict, the two candidate answers ``This cell is a \{type of A\}'' and ``This cell is a \{type of B\}'' are scored by per-token log-likelihood. The prediction is given by the answer with lower perplexity.

\subsection{Predictability of Underused Biological Representations}\label{app:rq1-probes}

\subsubsection{Genome-dependent disease prediction subset}\label{app:rq1-probe-dna}

Queries are grouped by the same prompt text (186 groups). A group is genome-dependent when it contains more than one distinct genome pair and more than one disease label. All other queries are genome-invariant, which means that every query with that text has the same label and a text-only predictor can score perfectly. The genome-dependent stratum has 165 queries in 14 text groups over 63 genomes and 12 labels, from six genes. With 126 train, 19 test and 20 validation queries. Within a group 118 queries carry the group's strictly most frequent label, 41 carry a less frequent label, and 6 queries belong to one RET group whose two labels tie.

\textbf{Evo2 probes.} We train linear probes on both the Evo2 representations as well as the projected Evo2 representations received by BioReason, $\mathbf{Z}_{\mathrm{Evo2}}$. In both, the reference and variant blocks are mean-pooled over their 2,048 positions and concatenated. The classifier is a multinomial
logistic regression with L2 regularisation on standardised features, without class weighting. We use five folds, grouped by genome, so a held-out variant's label must be inferred from other variants.

\textbf{Text-only predictor.} The text-only predictor is fitted inside the same cross-validation loop as the probe. For each fold it assigns to a held-out query the most frequent gold label of that query's text among the queries in the training fold. Averaged over the 10 fold assignments it scores 0.708 on the 165 queries.

\subsubsection{InterPro-controlled protein subset and functional targets}\label{app:rq1-probe-protein}

We use the 540 proteins from the evidence conflict across the three properties with 199 proteins in 31 InterPro families for enzyme activity, 232 in 87 InterPro families
for organelle targeting and 109 in 22 InterPro families for DNA binding, and 140 InterPro families in total. We use binary labels based on the presence of the GO term of the property \texttt{GO:0003824} or \texttt{GO:0008233} for enzyme activity, \texttt{GO:0005739} or \texttt{GO:0009507} for organelle targeting, and \texttt{GO:0003677} for DNA binding. A negative label means the curated GO annotation lacks the term. GO \texttt{NOT} pairs are not used because their readout terms differ between pairs, so no single label is shared across them.

We train the linear probe on predicting both the ESM3 representations and the projected representations that are used by BioReason-Pro. For each we use the mean-pooled representation over all residues. The classifier is a logistic regression with L2 regularisation at the library default on standardised features, fitted separately for the three properties, with five folds grouped by InterPro family. We score with AUROC within each InterPro family, and InterPro families are averaged with equal weight.

\subsection{Removing differentially expressed genes from cell sentences}\label{app:rq1-deg}

DEGs are computed per atlas, from the $\log(1+x)$ matrix with scanpy's Wilcoxon rank-sum test of each annotated type against all other cells. A gene is a DEG of a type when its $\log_2$ fold change is at least 1.0 and its Benjamini--Hochberg adjusted $p$-value is below 0.05. The fold change criterion is one-sided because the cell sentence is ranked by decreasing expression. DEGs are ranked by $\log_2$ fold change times $-\log_{10}$ of the adjusted $p$-value. For a given cell, its DEG set is the DEGs of its annotated type that occur in its own 1,000 gene sentence. The size of a cell's DEG set varies by atlas: the median is 82, 66, 55, 136 and 97 genes in immune1, immune2, immune3, pancreas and lung.

We remove proportions $f \in \{0.10, 0.25, 0.50, 0.75, 1.00\}$ of the cell's DEGs from the cell sentence. We compare four different ways of removing DEGs: (1) the highest ranked DEGs, (2) the weakest ranked DEGs (for $f \le 0.75$ because at $f = 1$ the set equals the set of strongest DEGs), (3) random genes drawn uniformly from the cell's own sentence, and (4) non-DEGs with similar expression given by the non-DEG of the same sentence with the nearest expression value in that cell to a DEG. Removed genes are deleted, not replaced, so the sentence shortens
by the same number of genes under all four conditions.

\subsection{Discussion of training-time availability of predictive text}\label{app:rq1-training}

Here we compare the six models for training-time availability of predictive text. For BioReason, each disease prediction query provides the wild type and variant DNA sequence representations followed by the chromosome, pathway definition, gene list with descriptions and the query naming the gene. No masking, dropout or augmentation of the inputs is applied. A majority-vote table from the gene symbol named in the query to the disease, fitted on the train split, results in test set accuracy of 0.932 across the 91 genes which all occur in training. Therefore, the text inputs that are always present at training time is sufficient for the task. The BioReason-Pro training script uses InterPro and the GO-GPT prediction. No dropout of either input is used. ChatNT's training queries give the task and the sequence representation. The query does not include details about the sequence, so no sequence-specific text can predict the label. Prot2Text-V2's released data loader replaces the protein name and the taxon each by \texttt{unknown} with probability 0.8, independently, at every training step. CellWhisperer's chat model is trained on conversations paired with a transcriptome representation, so the transcriptome representation is the input present in every prompt. C2S-Scale is trained to predict a cell type given only the cell sentence.

Across the six models, the two models with training prompts that always contain answer-predictive text (BioReason and BioReason-Pro) show nearly unchanged performance under sequence shuffling and prefer the text in evidence conflicts. Conversely, the four models trained without such text (ChatNT, CellWhisperer, C2S-Scale) or with high dropout rates (Prot2Text-V2) show reduced performance under biological input perturbations and, where evaluated, prefer the sequence or cell source in conflicts. We report this as an observation only. The models differ in architecture, task, training data and scale, so this comparison cannot isolate the effect of training prompt content, and we do not test it with a controlled intervention.

%% file: iclr2026/092appendix_rq2.tex
\section{RQ2: Post-training, model scale, and biological input contribution to performance}
\label{app:rq2}

\subsection{BioReason post-training sweep checkpoints}
\label{app:rq2-sweep}

The 42 checkpoints come from the post-training sweep of \citet{biopostrain} which includes the two final SFT and RL checkpoints. The language models are Qwen3-1.7B and Qwen3-4B. For each model size, a SFT sweep with LoRA rank $r \in \{16, 64, 256\}$ ($\alpha = 2r$, learning rate $5 \times 10^{-5}$, at most 4 epochs) and GRPO sweep with LoRA rank $r \in \{4, 16, 64\}$ (learning rate $10^{-5}$, 8 generations per prompt, 4 epochs or 1,156 steps) runs from each SFT run. The 42 checkpoints are 6 SFT and 36 GRPO checkpoints from 24 training runs. We use the same set up as defined in Appendix~\ref{app:rq1-perturb}.

\begin{figure}[h]
    \centering
    \includegraphics[width=0.68\linewidth]{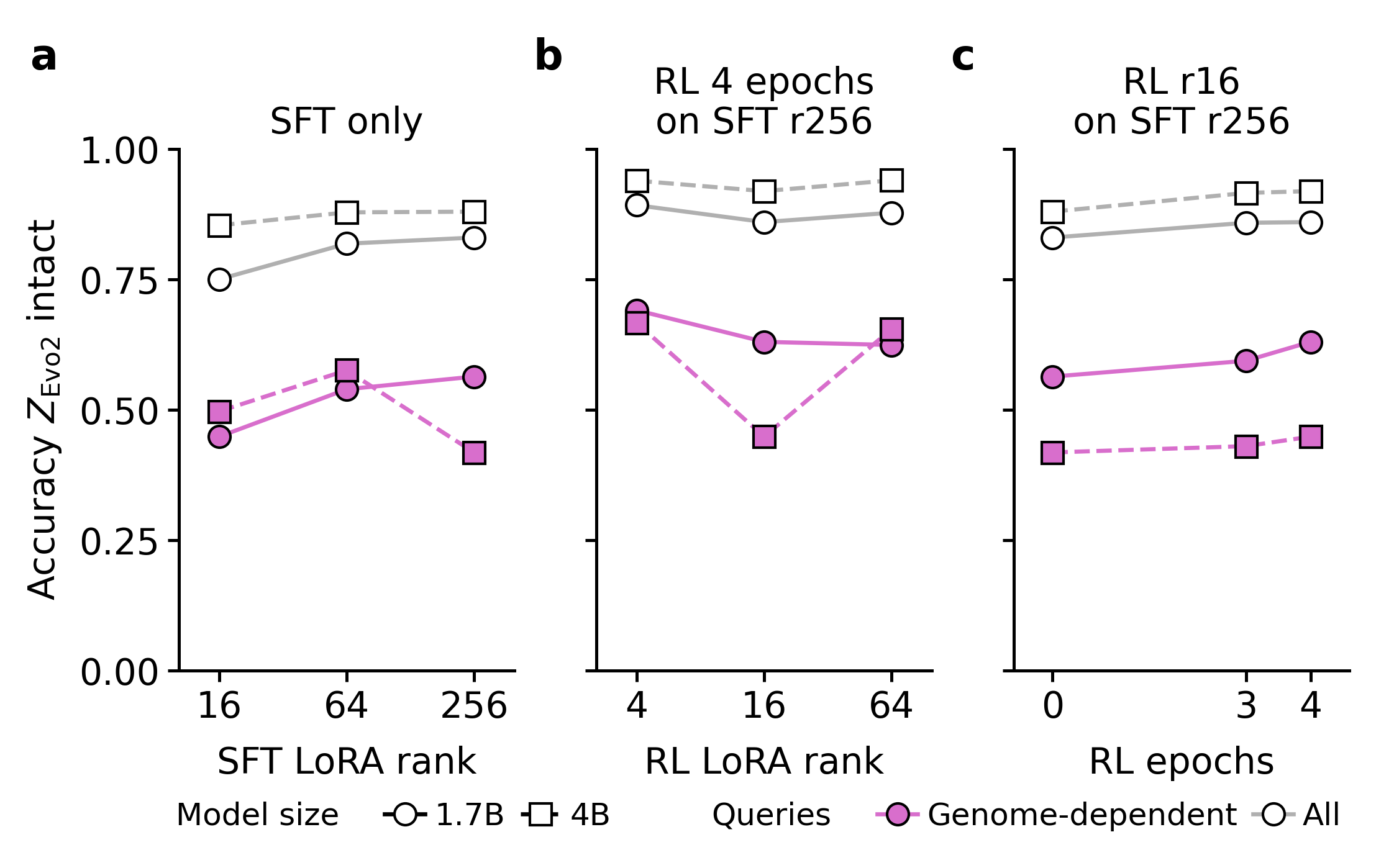}
    \caption{
    \textbf{Post-training changes BioReason performance without consistently increasing
    the performance contribution from $\mathbf{Z}_{\mathrm{Evo2}}$.}
    \textbf{a--c} Accuracy on the genome-dependent subset and all 1,449 queries as SFT LoRA
    rank, RL LoRA rank, or RL duration is varied.
    }
    \label{fig:posttraining_sweep}
\end{figure}

\paragraph{Across 42 BioReason checkpoints, post-training changes accuracy without consistently increasing the performance contribution from $\mathbf{Z}_{\mathrm{Evo2}}$.}
We broaden the comparison beyond the released SFT and RL checkpoints by evaluating 42 BioReason checkpoints \citep{biopostrain} that vary model size, SFT LoRA rank, RL LoRA rank, and RL duration (Fig.~\ref{fig:posttraining_sweep}). We focus on the 165 genome-dependent queries, where identical prompt text occurs with different genomes and different disease labels. Performance varies substantially across post-training configurations, but higher accuracy does not consistently coincide with a larger advantage for the original $\mathbf{Z}_{\mathrm{Evo2}}$ over the representation recomputed from shuffled DNA (Fig.~\ref{fig:sft_vs_rl}b).

This observation holds across individual post-training sweeps. Increasing SFT LoRA rank changes accuracy on the 165 queries from approximately 0.45--0.56 for the 1.7B model and 0.42--0.58 for the 4B model, while overall disease prediction accuracy increases to approximately 0.75--0.88 (Fig.~\ref{fig:posttraining_sweep}a). Varying RL LoRA rank from 4 to 64 or increasing RL duration from 0 to 4 epochs likewise changes both overall and genome-dependent accuracy (Fig.~\ref{fig:posttraining_sweep}b,c), without consistently increasing the advantage of the original genomic representation. Post-training can improve disease prediction accuracy, including on queries where the genome distinguishes the target label, without increasing the performance contribution from $\mathbf{Z}_{\mathrm{Evo2}}$.

\subsection{C2S-Scale 2B and 27B}
\label{app:rq2-c2s}

Both checkpoints are evaluated on the same 4,846 cells from RQ1. Both models receive the same questions with the same set up outlined in Appendix~\ref{app:rq1-deg}.

%% file: iclr2026/093appendix_rq3.tex
\section{RQ3: Reasoning traces and rationales}
\label{app:rq3}
\subsection{BioReason reasoning traces}

We evaluate the released RL and SFT checkpoints on the 145 queries held out by the pathway network split. Each query is generated three times, with the original reference and variant windows, with the identical prompt a second time, and with both windows shuffled. Decoding is deterministic for both checkpoints, so any difference under shuffling follows from the input rather than from sampling. Of the 145 queries.

xWe use deterministic string matching to extract the stated base change, coordinate, chromosome, and gene from each reasoning trace, retaining the first match for each claim type.
Accuracy on the stated base pair is reported over the 111 queries whose variant is a single base substitution, since a single base pair change cannot match a deletion, an insertion or a multi-base change. The coordinate in the reasoning trace is compared with the position given in the dataset, which is available for 137 of the 145 queries. The input prompt provides the chromosome number and the two 2,049 bp windows and absolute position is not provided.

\subsection{BioReason-Pro reasoning traces}

We use the evidence conflicts of Appendix~\ref{app:rq1-conflict} in the direction that takes the InterPro from protein $B$ and uses protein $A$ for the $\mathbf{Z}_{\mathrm{ESM3}}$ and the GO-GPT input, on the 189 pairs where the model recovers both members' property from their own unperturbed prompts. We keep the evidence conflict pairs by mapping the InterPro accessions through the EMBL-EBI \texttt{interpro2go} file and taking the \texttt{is\_a} and \texttt{part\_of} closure of the property term in the Gene Ontology.
The reasoning model is not given this mapping from InterPro to GO terms. The InterPro accession of protein $B$ corresponds the evaluated GO term on 66 of the 189 pairs, and on 53 of those neither InterPro accessions nor its GO-GPT output of the protein $A$ contains the GO term.

We use deterministic string matching to extract the InterPro accessions it references in the reasoning trace by matching on \texttt{IPR} followed by six digits. This reference InterPro accessions is equal to the input InterPro accessions from protein $B$ on 53 of 53 queries, so quoting the reasoning model always references the accession that maps to the property protein $A$ does not have. We also use string matching to determine whether a sentence quoting that accession states the property it implies. We delete every InterPro entry name printed in the prompt from the sentence and apply one fixed pattern per property term. A match is kept only if the matched phrase is not part of any printed entry name.
The property of protein $B$ is therefore stated in 23 of 53 reasoning traces. However, all 53 queries omit the GO term this property implies in the final response.

\subsection{Cell type annotation rationales with Cell2Sentence}
\subsubsection{Rationale generation}
\label{app:rq3-generation}
We sample 400 cells per atlas from the 4,846 annotation cells of Table~\ref{tab:models_tasks}. We use the same DEG perturbations as defined in Appendix~\ref{app:rq1-deg}: the unmodified sentence of the 1,000 most highly expressed genes in descending order, removal of the strongest DEGs of the annotated type at 50\% and 100\%, removal of the same number of non-DEGs with the closest expression value at 50\% and 100\%, and the unmodified gene set under one uniform random permutation per cell so the input position is independent expression. We include the random permutation to determine whether genes cited in the rationale are preferentially selected because of their expression level or their position in the expression-ranked input.

Every prompt has the form
\begin{quote}\ttfamily\small\raggedright
The following is a list of \{n\} gene names ordered by descending expression level in a Homo
sapiens cell. Your task is to give the cell type which this cell belongs to based on its gene
expression.\\
Cell sentence: \{genes\}.\\
Answer using exactly this format:\\
Cell type: \textless{}cell type\textgreater{}\\
Rationale: \textless{}the genes that support the call\textgreater{}\\
\mbox{}\\
Cell type:
\end{quote}
where \texttt{\{genes\}} is the space-separated gene sentence and \texttt{\{n\}} its length.

For C2S-Scale 2B with the same template, decoding settings, and conditions, no generated response contains the string \texttt{Rationale}. Conversely, for the 27B checkpoint, 12,000 generations (2,000 cells $\times$ six conditions) contain the \texttt{Rationale:} field. RQ3 therefore reports the 27B checkpoint only.

\subsubsection{Gene extraction and analysis of referenced genes}
\label{app:rq3-extraction}

We extract referenced genes from the full generation. We split the text into tokens that start with a letter or digit and continue with letters, digits, hyphens, or periods, upper-case each token, remove one trailing period, and keep the tokens that are in the vocabulary. Repeated mentions count once, at their first occurrence, and the order of first occurrences defines the mention order used for first-$k$ precision.

\begin{figure}[h]
    \centering
    \includegraphics[width=0.8\linewidth]{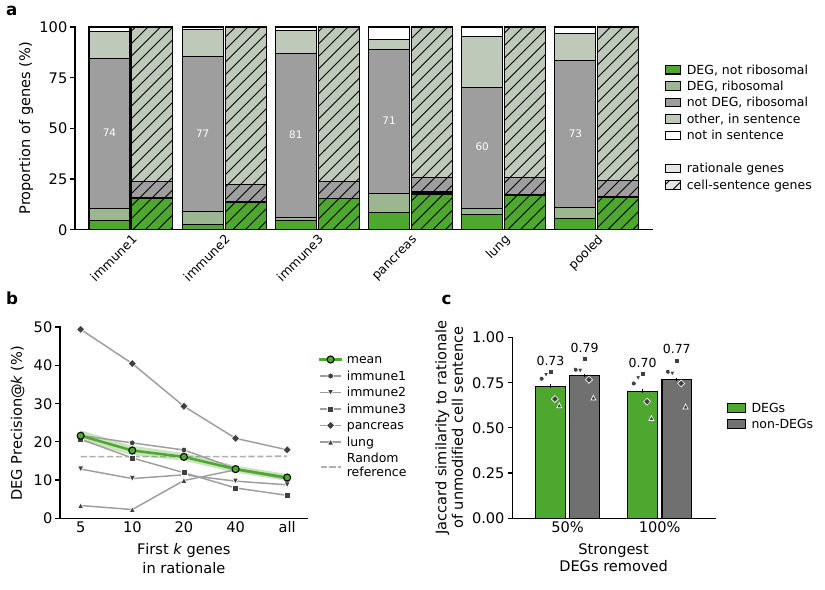}
    \caption{
    \textbf{Genes referenced in C2S-Scale rationales, and how that set changes when DEGs are
    removed from the cell sentence.}
    \textbf{a} Composition of referenced genes across five atlases.
    \textbf{b} DEG precision among the first $k$ referenced genes.
    \textbf{c} Jaccard similarity to the unmodified set of referenced genes after removing the
    strongest DEGs or non-DEGs with similar expression from $\mathbf{Z}_{\mathrm{cell}}$.
    Error bars are 95\% CIs.
    }
    \label{fig:rq3_rationale_supp}
\end{figure}

For a cell with annotated type $t$, let $S$ be the genes of the input cell sentence, $D_t$ the DEGs of $t$, and $g_1, g_2, \ldots$ the referenced genes in order of mention with $G_k = \{g_1, \ldots, g_k\}$. DEG precision among the first $k$ referenced genes is given by,
\[
\mathrm{Prec}@k = \frac{\sum_{c} |G_k \cap D_t \cap S|}{\sum_{c} |G_k|}.
\]
The random reference, the dashed line in Fig.~\ref{fig:rq3_rationale_supp}b
and the hatched bars in Fig.~\ref{fig:rq3_traces}d, is the expected precision of the same number of genes drawn uniformly from $S$.

Let $G$ and $G'$ be the sets of referenced genes under the unmodified and perturbed sentences and $R$ the set of genes removed from the input. Fig.~\ref{fig:rq3_rationale_supp}c reports the Jaccard similarity $|G \cap G'| / |G \cup G'|$, computed per cell and averaged over cells.

\textbf{Ribosomal genes.} Genes are annotated as ribosomal protein genes by membership in HGNC gene groups 646, 728, and 729 (L, S, and mitochondrial ribosomal proteins) this results in a total of 171 approved symbols. Fig.~\ref{fig:rq3_rationale_supp}a assigns each referenced gene to one category in the order (1) absent from $S$, (2) DEG of $t$ in $S$, split into ribosomal and non-ribosomal, (3) ribosomal non-DEG in $S$, and (4) other gene in $S$. The proportion of ribosomal counts every referenced ribosomal protein gene,
including ribosomal DEGs and ribosomal genes absent from $S$. The cell sentence bars of Fig.~\ref{fig:rq3_rationale_supp}a are the expected composition of a same-size uniform draw from $S$.

\subsubsection{Instructing the rationale to name DEGs}
\label{app:rq3-degprompt}
We further test whether their DEG precision depends on that query by testing seven queries that ask for marker genes or DEGs, on 200 cells (40 per atlas). Each
cell is compared with its own generation under the unmodified condition. On these 200 cells the original query gives a DEG precision of 11.2\%, a precision among
the first five referenced genes of 23.2\%, and a ribosomal share of 77.9\%. The seven queries replace the answer format ending, and where it is stated with respect to the cell sentence.
\begin{description}[leftmargin=0pt,itemsep=1pt]
\item[Request for marker genes.] \texttt{<genes from the cell sentence that are specifically
expressed in this cell type and distinguish it from other cell types>}.
\item[Request for differentially expressed genes.] \texttt{<genes from the cell sentence that
are more highly expressed in this cell type than in other cell types>}.
\item[Request for marker genes with exclusions.] \texttt{<up to 10 genes from the cell sentence that are specific to this cell type; do not list ribosomal, mitochondrial or housekeeping genes>}.
\item[Same as above, instruction before the sentence.] The same request placed in the task sentence rather than the answer format.
\item[Annotated cell type given, request for marker genes only.] No cell type is generated. The prompt states the atlas label and asks for its marker genes, with the same exclusions of ribosomal, mitochondrial or housekeeping genes.
\item[Predicted cell type given, request for marker genes only.] As above, with the type the model produced for that cell under the unmodified condition in place of the atlas label.
\item[One-shot learning.] An \texttt{Example:} block shows another cell of the same atlas with a different annotated type, its sentence cut to 200 genes, its type, and its ten strongest DEGs as marker genes, followed by the evaluated cell under \texttt{Now the cell to annotate:}.

\end{description}

\begin{table}[h]
\centering
\caption{\textbf{Variation on the rationale query and DEG precision of C2S-Scale 27B.} Referenced genes per generation, ribosomal protein share, DEG precision against the annotated type (random draw in parentheses), precision among the first five referenced genes (Prec.@5), and cell type accuracy, pooled over 200 cells (40 per atlas). Accuracy is undefined when the type is supplied in the prompt.}
\label{tab:rq3-degprompt}
\footnotesize
\setlength{\tabcolsep}{4pt}
\begin{tabular}{@{}>{\raggedright\arraybackslash}p{6cm}rrrrr@{}}
\toprule
Query & Genes & Ribo.\ (\%) & DEG prec.\ (\%) & Prec.@5 (\%) & Acc. \\
\midrule
original query & 66.5 & 77.9 & 11.2 (16.2) & 23.2 (16.1) & 0.40 \\
\midrule
Request for marker genes & 62.5 & 72.6 & 12.0 (16.4) & 22.3 (16.3) & 0.41 \\
Request for differentially expressed genes & 48.4 & 71.3 & 13.2 (16.3) & 20.6 (16.4) & 0.42 \\
Request for marker genes with exclusions & 61.6 & 69.9 & 13.3 (16.3) & 21.8 (16.3) & 0.42 \\
Same as above, instruction before the sentence & 65.8 & 74.9 & 11.6 (16.3) & 23.9 (16.1) & 0.40 \\
Annotated cell type given, request for marker genes only & 65.5 & 69.8 & 13.3 (16.2) & 20.9 (16.1) & -- \\
Predicted cell type given, request for markers only only & 65.0 & 69.0 & 12.2 (16.3) & 20.1 (16.1) & -- \\
One-shot learning & 57.0 & 64.5 & 10.4 (16.1) & 14.5 (16.2) & 0.18 \\
\bottomrule
\end{tabular}
\end{table}

Pooled DEG precision is 10.4 to 13.3\% across the seven queries, compared to 11.2\% for the original query on the same cells while DEGs of the annotated  cell type comprise 16.2\% of the genes in the cell sentence. The largest paired increase is 2.1 points for the query that requests at most ten marker genes and excludes ribosomal, mitochondrial, and housekeeping genes. This also increases precision among the first five referenced genes by $-1.4$ points. One-shot learning also lowers accuracy from 0.40 to 0.18.
\subsubsection{Rationale content by prediction correctness}
\label{app:rq3-stratified}
We split the rationales for the unmodified cell sentences by whether the predicted cell type is correct and compute DEG precision against the DEGs of the annotated type and of the predicted type. A prediction is correct when the generated label and the annotated type resolve to the same Cell Ontology term. The predicted type is the atlas type named by the generated label and is defined for 1,200 of the 2,000 cells. We report the same quantities for the prompt with cell type options of Appendix~\ref{app:rq3-label-change}. The prompt that supplies the annotated type and requests only marker genes (Appendix~\ref{app:rq3-degprompt}) generates no label and is reported as a separate condition.

\begin{table}[h]
\centering
\caption{\textbf{DEG precision of C2S-Scale 27B rationales by prediction correctness.} Precision among the first five referenced genes (Prec.@5) and among all referenced genes (Prec.@all), in \%, with the precision of a uniform random draw of the same number of genes from the cell sentence in parentheses. Annotated and predicted columns use the DEGs of the annotated and the predicted cell type. Pooled over five atlases. -- marks rows that include cells whose label names no atlas type.}
\label{tab:rq3-stratified}
\footnotesize
\setlength{\tabcolsep}{4pt}
\begin{tabular}{@{}lrcccc@{}}
\toprule
 & & \multicolumn{2}{c}{Annotated type} & \multicolumn{2}{c}{Predicted type} \\
\cmidrule(lr){3-4}\cmidrule(l){5-6}
Cells & $N$ & Prec.@5 & Prec.@all & Prec.@5 & Prec.@all \\
\midrule
\multicolumn{6}{@{}l}{\textit{Original prompt}} \\
All & 2,000 & 21.6 (16.1) & 10.6 (16.2) & -- & -- \\
Correct prediction & 748 & 35.2 (19.9) & 16.9 (20.0) & 35.4 (20.6) & 17.4 (20.7) \\
Incorrect prediction & 1,252 & 13.4 (13.8) & \phantom{0}6.9 (13.9) & -- & -- \\
\quad predicted type in atlas & 452 & 13.7 (\phantom{0}9.4) & \phantom{0}5.0 (\phantom{0}9.5) & 24.2 (\phantom{0}9.9) & \phantom{0}8.1 (10.1) \\
\midrule
\multicolumn{6}{@{}l}{\textit{Prompt with cell type options}} \\
All & 2,000 & 20.5 (16.1) & 10.7 (16.2) & -- & -- \\
Correct prediction & 821 & 33.1 (18.3) & 15.2 (18.4) & 33.6 (20.0) & 16.4 (20.0) \\
Incorrect prediction & 1,179 & 11.7 (14.5) & \phantom{0}7.5 (14.6) & -- & -- \\
\quad predicted type in atlas & 622 & 13.7 (10.1) & \phantom{0}5.7 (10.0) & 14.9 (\phantom{0}7.8) & \phantom{0}8.6 (\phantom{0}7.8) \\
\bottomrule
\end{tabular}
\end{table}

For C2S-Scale 27B with the original prompt, precision among the first five referenced genes is 35.2\% for correct predictions against 19.9\% for the random draw, and 13.4\% for incorrect predictions against 13.8\%. For the 452 incorrect predictions whose label names an atlas type, precision among the first five genes is 24.2\% against the DEGs of the predicted type and 13.7\% against those of the annotated type, with random draws of 9.9\% and 9.4\%.

\subsubsection{Effect of the rationale request on the predicted label}
\label{app:rq3-label-change}
The rationale prompt of Appendix~\ref{app:rq3-generation} has no option list and a different answer format from the multiple-choice prompt of Appendix~\ref{app:models}. For the same 2,000 cells we generate two responses from prompts that share the task sentence, the option list in the order of the multiple-choice evaluation, the cell sentence, and the answer format, and differ only in the \texttt{Rationale:} line:
\begin{quote}\ttfamily\small\raggedright
Answer using exactly this format:\\
Cell type: \textless{}cell type, copied verbatim from the list above\textgreater{}\\
\textit{[Rationale: \textless{}the genes that support the call\textgreater{}]}\\
\mbox{}\\
Cell type:
\end{quote}
The label is the first line of the generation. Accuracy uses the Cell Ontology rule of Appendix~\ref{app:rq3-stratified}, because the pancreas and lung options are study labels such as \texttt{alpha} and the model answers with Cell Ontology names under both prompts.

\begin{table}[h]
\centering
\caption{\textbf{Change in the C2S-Scale 27B predicted label when the rationale is requested.} Share of cells whose label string or resolved Cell Ontology type differs between the two prompts, cell type accuracy without and with the rationale request, and the number of cells whose prediction changes from correct to incorrect and from incorrect to correct. 400 cells per atlas.}
\label{tab:rq3-label-change}
\footnotesize
\setlength{\tabcolsep}{4pt}
\begin{tabular}{@{}lrrrrrr@{}}
\toprule
 & \multicolumn{2}{c}{Changed (\%)} & \multicolumn{2}{c}{Accuracy} & \multicolumn{2}{c}{Cells} \\
\cmidrule(lr){2-3}\cmidrule(lr){4-5}\cmidrule(l){6-7}
Atlas & Label & Type & w/o rationale & w/ rationale & Correct to incorrect & Incorrect to correct \\
\midrule
immune1 & 21.5 & 9.8 & 0.71 & 0.61 & 49 & 7 \\
immune2 & 62.5 & 44.0 & 0.12 & 0.08 & 27 & 8 \\
immune3 & 16.0 & 9.8 & 0.53 & 0.44 & 37 & 3 \\
pancreas & 18.0 & 9.8 & 0.65 & 0.60 & 22 & 2 \\
lung & 34.5 & 18.0 & 0.26 & 0.34 & 5 & 37 \\
\midrule
Pooled & 30.5 & 18.2 & 0.45 & 0.41 & 140 & 57 \\
\bottomrule
\end{tabular}
\end{table}

For C2S-Scale 27B, adding the rationale request changes the label string for 30.5\% of the 2,000 cells (95\% CI 28.5--32.5) and the resolved cell type for 18.2\% (95\% CI 16.6--19.9). Pooled accuracy decreases from 0.45 to 0.41, a difference of $-4.2$ points (95\% CI $-5.5$ to $-2.8$). Accuracy decreases in four atlases and increases from 0.26 to 0.34 in lung, where the four basal subtypes resolve to one Cell Ontology term.

%% file: iclr2026/094appendix_rq4.tex
\section{RQ4: Auxiliary sequence supervision}
\label{app:rq4}

\subsection{Data}
\label{app:rq4-data}

We use the pathway network split of the 1,449 KEGG queries from the BioReason post-training code base \citep{biopostrain} which includes 1,159 training, 145 validation, and 145 held-out queries. Since the split holds out pathway networks, in Table~\ref{tab:rq4-overlap} we report the overlap for other properties with the held-out set. We use genome, as in Table~\ref{tab:models_tasks}, for a distinct pair of reference and variant windows. For 135 of the 145 held-out queries the same genome occurs in training under a different prompt, and 133 of these have the same disease label there. The held-out queries cover 8 of the 37 disease labels with 60 of them
labeled amyotrophic lateral sclerosis. The mutations are 111 single-base substitutions, 20 deletions, 9 insertions, and 5 multi-base substitutions.

\begin{table}[h]
\caption{Overlap between the 145 held-out queries and the 1,159 training queries.}
\label{tab:rq4-overlap}
\centering
\small
\begin{tabular}{lrrr}
\toprule
\textbf{Held-out property} & \textbf{Distinct values} & \textbf{Also in training} & \textbf{Queries with the value in training} \\
\midrule
Prompt text                & 12  & 0  & 0 / 145 \\
Network definition         & 9   & 0  & 0 / 145 \\
Gene named in the query    & 8   & 7  & 139 / 145 \\
Genome (both windows)      & 100 & 90 & 135 / 145 \\
Disease label              & 8   & 8  & 145 / 145 \\
\bottomrule
\end{tabular}
\end{table}

\subsection{Auxiliary target and training}
\label{app:rq4-training}

Both windows of a query start at the same coordinates with 128~bp either side for the 257-bp condition and 1,024~bp before and 1,023~bp after for the 2,048-bp condition, so the edit
sits at the same fixed offset for training and evaluation. For an insertion or
deletion the two cuts agree before the edit and are shifted relative to each other after it. The auxiliary target replaces the reasoning trace with one line, \texttt{Edit at $i$: ref X -> var Y (class)}, where $i$ is the offset of the edit, X and Y are the reference and variant bases there, and the class is substitution, insertion, deletion, or multi-base. The control keeps the benchmark's reasoning trace, which has a median of 2,197 characters and names the substituted bases in 865 of the 1,159 training queries. The bases are available only from the input DNA representations. One training query of the 257-bp run is

\begin{footnotesize}
\begin{verbatim}
<|im_start|>user
<|dna_start|>[257 slots, reference]<|dna_end|>
<|dna_start|>[257 slots, variant]<|dna_end|>Chromosome Number: 16
Network Definition of the pathway: mRNA(pre) -- FUS* // mRNA
Genes in the pathway: FUS; FUS RNA binding protein

Given this context, what is the biological effect of this FUS allele,
specifically what disease does this contribute to?<|im_end|>
<|im_start|>assistant
<think>
Edit at 128: ref A -> var T (substitution)
</think>

Answer: amyotrophic lateral sclerosis<|im_end|>
\end{verbatim}
\end{footnotesize}

We use the same training set up as \cite{biopostrain}. The text model is the Qwen3-4B checkpoint after continued pretraining used by \citet{biopostrain}, adapted with LoRA (rank 32, $\alpha=64$) on every linear layer except the output head, which trains 66.1M parameters. The loss is the next-token cross-entropy over the assistant turn plus five times the cross-entropy over the tokens after a marker string, \texttt{Edit at} in the auxiliary runs and \texttt{Answer:} in the control, so the weighted term includes the edited bases and the disease in
the auxiliary runs and the disease alone in the control. All runs use AdamW at learning rate $5\times10^{-5}$, batch size 1 with 8 gradient accumulation steps, and 12 epochs, on one H100 per run. We train three independent seeds and evaluate the checkpoints from each training run with the lowest validation loss.

\subsection{Evaluation and results}
\label{app:rq4-results}

Each model generates for every held-out query under three conditions with the prompt fixed: (1) the intact DNA sequences; (2) both DNA sequences shuffled which
changes $\mathbf{Z}_{\mathrm{Evo2}}$; and (3) an evidence conflict that replaces both windows with those of another held-out query drawn at random. Accuracy on the edited base pair is determined by the first reference-to-variant base pair stated in the generated reasoning, which in the auxiliary runs is the pair in the \texttt{Edit at} line. The
text-only reference uses no DNA and uses the most frequent training disease for the gene named in the prompt which reaches 0.917, and always writing the most frequent training variant pair reaches 0.103. We report the results in Table~\ref{tab:rq4-results}.

\begin{table}[h]
\caption{Accuracy on the 145 held-out queries, as the mean over three training seeds and the standard
error of that mean. $\Delta$ is intact minus shuffled. The conflict column is the share of queries
whose written base pair equals the donor's.}
\label{tab:rq4-results}
\centering
\small
\setlength{\tabcolsep}{4pt}
\begin{tabular}{lccccc}
\toprule
& \multicolumn{2}{c}{\textbf{Disease}} & \multicolumn{2}{c}{\textbf{Edited base pair}} & \textbf{Conflict} \\
\cmidrule(lr){2-3}\cmidrule(lr){4-5}
\textbf{Training target, window} & \textbf{Intact} & \textbf{$\Delta$} & \textbf{Intact} & \textbf{$\Delta$} & \textbf{Donor's pair} \\
\midrule
No auxiliary, 2,048~bp & 0.901 (0.026) & 0.002 (0.022) & 0.080 (0.002) & 0.005 (0.002) & 0.071 (0.005) \\
Auxiliary, 257~bp      & 0.970 (0.009) & 0.021 (0.007) & 0.671 (0.041) & 0.549 (0.032) & 0.467 (0.072) \\
Auxiliary, 2,048~bp    & 0.984 (0.005) & 0.028 (0.007) & 0.531 (0.115) & 0.432 (0.123) & 0.361 (0.139) \\
\midrule
Text-only reference    & 0.917 & & 0.103 & & -- \\
\bottomrule
\end{tabular}
\end{table}

We split accuracy on the edited base pair by whether the query's genome occurs in training, over the 145 held-out and 145 validation queries (Table~\ref{tab:rq4-unseen}). Only 72 of the 290 queries, 10 held-out and 62 validation, have a genome absent from training.

\begin{table}[h]
\caption{Accuracy on the edited base pair for held-out and validation queries whose genome is absent from
(72 queries) or present in (218 queries) the training set, as the mean over three training seeds.}
\label{tab:rq4-unseen}
\centering
\small
\begin{tabular}{lcc}
\toprule
\textbf{Training target, window} & \textbf{Unseen genomes} & \textbf{Seen genomes} \\
\midrule
No auxiliary, 2,048~bp & 0.042 & 0.083 \\
Auxiliary, 257~bp      & 0.148 & 0.691 \\
Auxiliary, 2,048~bp    & 0.106 & 0.606 \\
\midrule
Most frequent training pair (T$>$C) & 0.111 & 0.101 \\
\bottomrule
\end{tabular}
\end{table}

\begin{figure}[t]
    \centering
    \includegraphics[width=0.8\linewidth]{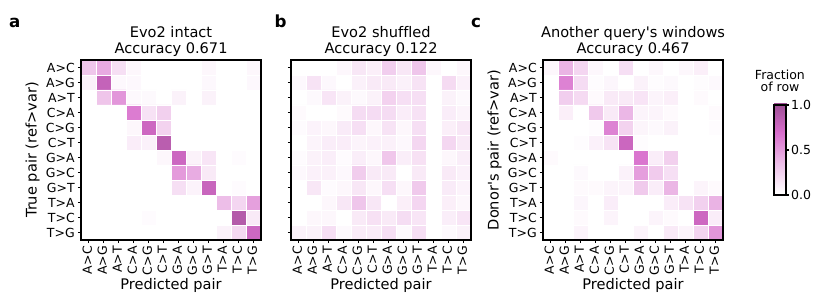}
    \caption{
    \textbf{Predicted reference-to-variant base pairs for the 257 bp auxiliary condition.}
    \textbf{a} Original DNA, \textbf{b} shuffled DNA, and \textbf{c} an evidence conflict that
    replaces both windows with those of another held-out query. Rows are the query's true pair in
    \textbf{a} and \textbf{b} and the donor's pair in \textbf{c}. Each row is normalized by its own
    total, so a diagonal cell is the accuracy for that pair. The three training seeds are pooled,
    giving $145\times3=435$ generations per matrix.
    }
    \label{fig:rq4_auxiliary_matrices}
\end{figure}